\documentclass[10pt,twocolumn,letterpaper]{article}

\usepackage{cvpr}              

\usepackage[dvipsnames]{xcolor}

\newcommand{\Fig}[1] {Fig.~\ref{fig:#1}}
\newcommand{\Figs}[1]{Figs.~\ref{fig:#1}}

\newcommand{\Tbls}[1] {Tables~\ref{tbl:#1}}
\newcommand{\Sec}[1] {Section~\ref{sec:#1}}

\usepackage{cite}
\usepackage{graphicx}
\usepackage{fmtcount}
\usepackage{epsfig} 
\usepackage{amsmath}
\usepackage{amssymb}  
\usepackage{booktabs} 
\usepackage{subcaption} 
\usepackage{lipsum}
\usepackage{arydshln} 
\usepackage{comment}
\usepackage[cjk]{kotex}

\definecolor{cvprblue}{rgb}{0.21,0.49,0.74}
\usepackage[pagebackref,breaklinks,colorlinks,citecolor=cvprblue]{hyperref}

\def\paperID{*****} 
\def\confName{3DV\xspace}
\def\confYear{2025\xspace}

\title{Deep Polycuboid Fitting for Compact 3D Representation of Indoor Scenes}

\author{
Gahye Lee \quad\quad  Hyejeong Yoon \quad\quad  Jungeon Kim \quad\quad  Seungyong Lee \\ 
POSTECH\\  
{\tt\small \{gahye0509, hjyoon02, jungeonkim, leesy\}@postech.ac.kr} 
}

\begin{document}
\maketitle
\begin{abstract}

This paper presents a novel framework for compactly representing a 3D indoor scene using a set of polycuboids through a deep learning-based fitting method. 
Indoor scenes mainly consist of man-made objects, such as furniture, which often exhibit rectilinear geometry.
This property allows indoor scenes to be represented using combinations of polycuboids, providing a compact representation that benefits downstream applications like furniture rearrangement. 
Our framework takes a noisy point cloud as input and first detects six types of cuboid faces using a transformer network. Then, a graph neural network is used to validate the spatial relationships of the detected faces to form potential polycuboids. 
Finally, each polycuboid instance is reconstructed by forming a set of boxes based on the aggregated face labels.
To train our networks, we introduce a synthetic dataset encompassing a diverse range of cuboid and polycuboid shapes that reflect the characteristics of indoor scenes.
Our framework generalizes well to real-world indoor scene datasets, including Replica, ScanNet, and scenes captured with an iPhone. 
The versatility of our method is demonstrated through practical applications, such as virtual room tours and scene editing. 
  
\end{abstract}     
\section{Introduction}
\label{sec:intro}

\begin{figure}[t]  
\begin{center}
  \includegraphics[width=\linewidth]{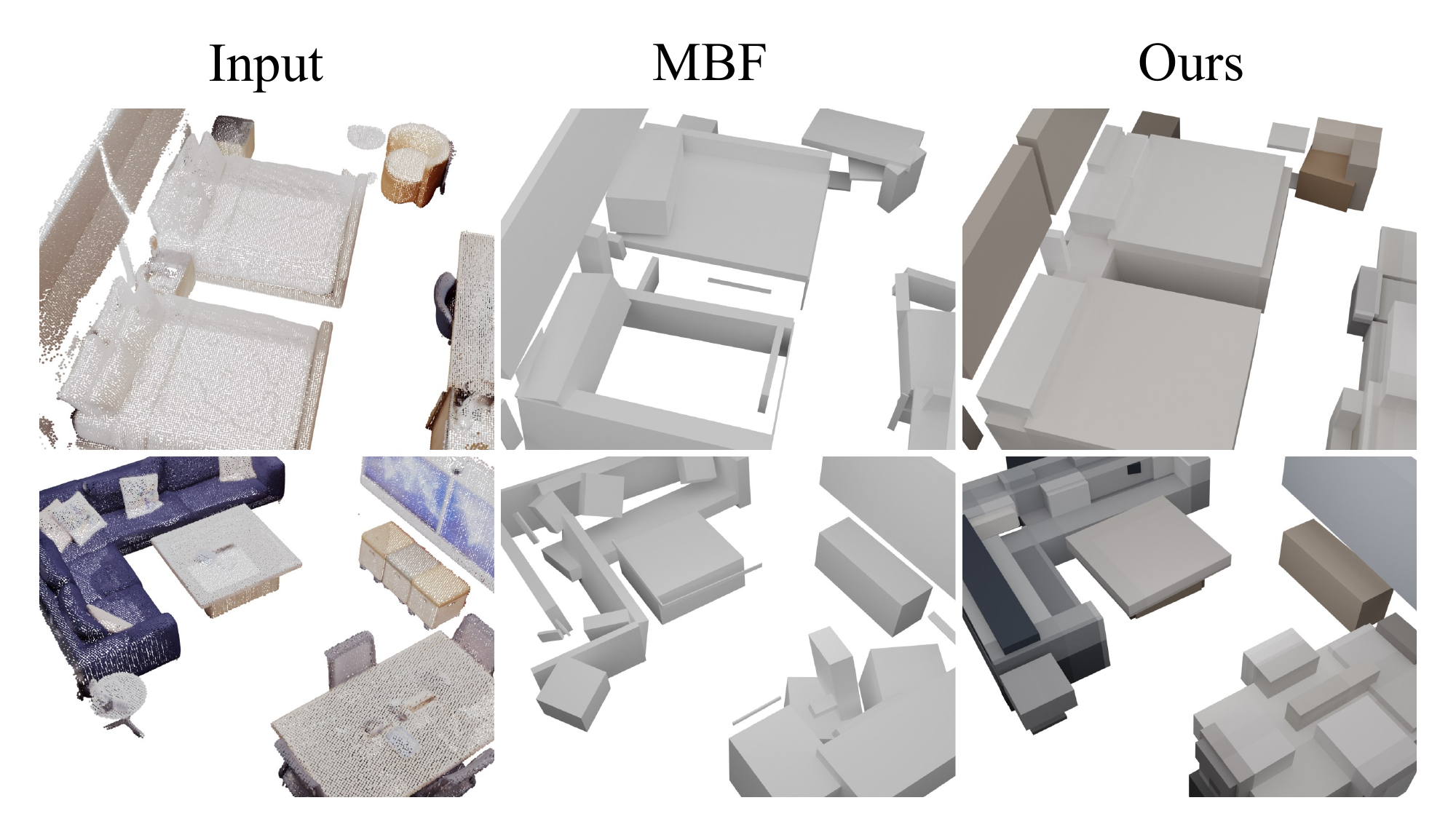}
\end{center} 
\caption{Comparison of shape abstraction results on Replica dataset. 
Compared to MBF \cite{ramamonjisoa2022monteboxfinder}, which generates many thin boxes covering only small portions of input points, our method detects high-quality polycuboids that comprehensively abstract the overall shapes of objects. 
}
\label{fig:quali_replica_zoomin}
\end{figure}

Understanding 3D scene geometry is a fundamental challenge in computer vision. 
While point clouds have been widely utilized for representing 3D scene geometry, they primarily serve as a raw data representation that inherently lacks part-level and object-level structures.
To address this limitation, shape abstraction approaches have been developed to model 3D geometries as compositions of simple primitives, such as planes [7,34], cuboids [5,36], and superquadrics [14,16]. 
These abstract representations provide a structured and interpretable description of scene geometry, which is useful for various applications, such as scene understanding~ \cite{shao2014imagining, lin2013holistic}, robot navigation~\cite{ruan2022efficient,ruan2022collision}, scene reconstruction \cite{huang20173dlite, wang2019efficientplane}, and AR/MR~\cite{chekhlov2007ninja,kim2022integrating}.
 
 

Indoor scenes consist of interior surfaces, such as walls, floors, and ceilings, as well as man-made objects including furniture and appliances. 
Given that these components often exhibit rectilinear shapes, cuboids could be a preferred representation for indoor scenes~\cite{yang2021unsupervised, sun2019learning, tulsiani2017learning}. 
However, cuboids inherently struggle to represent concave structures, such as L-shaped sofas, and are insufficient to capture the diversity of object shapes in indoor scenes.
In contrast, polycuboids, which combine multiple cuboids, can provide a more expressive and accurate representation.

In this paper, we explore the use of polycuboids to compactly represent the underlying geometries of 3D indoor scenes. Specifically, we introduce a data-driven approach that learns to fit a set of polycuboids to a noisy point cloud. 
To the best of our knowledge, leveraging polycuboids for 3D representation has not been explored previously. 

Fitting polycuboid instances to a noisy point cloud is challenging. 
Constructing reliable priors that generalize across diverse polycuboid configurations and object shapes is inherently ill-posed.
Partial or noisy scan data further complicates the robust detection of polycuboids.
Moreover, existing cuboid fitting methods cannot be directly extended to construct polycuboids, as they typically rely on heuristic plane fitting algorithms and tend to overly decompose scenes (\Fig{quali_replica_zoomin}). 
In that case, additional filtering or merging steps are required to refine small or thin cuboids and obtain a structured and meaningful polycuboid representation.

In this paper, we explore the geometric priors of a cuboid, the fundamental unit of a polycuboid, for polycuboid detection,
We define six cuboid face types and twelve spatial relationships, encoded using discrete and finite labels.
By extending these priors, we adapt cuboid-based relationships to comprehensively represent diverse structures of polycuboids (\Fig{cube system}).
Instead of a top-down approach, we adopt a bottom-up approach detecting individual polycuboid faces and then aggregating them into polycuboid instances. 

Given a point cloud, we detect individual polycuboid faces using a transformer network.
A graph neural network is then used to infer spatial relationships of the detected faces to form potential polycuboid instances. 
Finally, each polycuboid instance is reconstructed by assembling cuboids based on the label information of aggregated faces. 
Our method enables robust detection of plausible polycuboids from a noisy or even incomplete point cloud (\Fig{quali_replica_zoomin}).
 
To train our transformer and graph neural networks, we introduce a synthetic polycuboid dataset encompassing a diverse range of cuboid and polycuboid shapes that reflect the characteristics of indoor scenes.  
Leveraging this synthetic dataset, our framework can faithfully handle the diverse shapes in indoor scenes.
We evaluate our method on three types of real scan dataset, ScanNet \cite{dai2017scannet}, Replica \cite{replica19arxiv}, and scenes captured using an iPhone. 
We also explore practical applications of our framework, such as virtual room tour and scene editing, combined with an off-the-shelf texture mapping technique.
 
Our main contributions can be summarized as follows;  
\begin{itemize}
\item We introduce the polycuboid representation for representing individual components of an indoor scene in a lightweight and controllable manner. 
\item We present a novel deep learning-based framework for polycuboid fitting of a noisy input point cloud, formulating the problem as cuboid component labeling and their aggregation into polycuboid instances.
\item We demonstrate the effectiveness of our framework on real-world indoor scene datasets, and show its practical applications such as virtual room tours and scene editing. 

\end{itemize}

\begin{figure*}
\begin{center}
\includegraphics[width=\linewidth, trim=0cm 4cm 1cm 4cm]{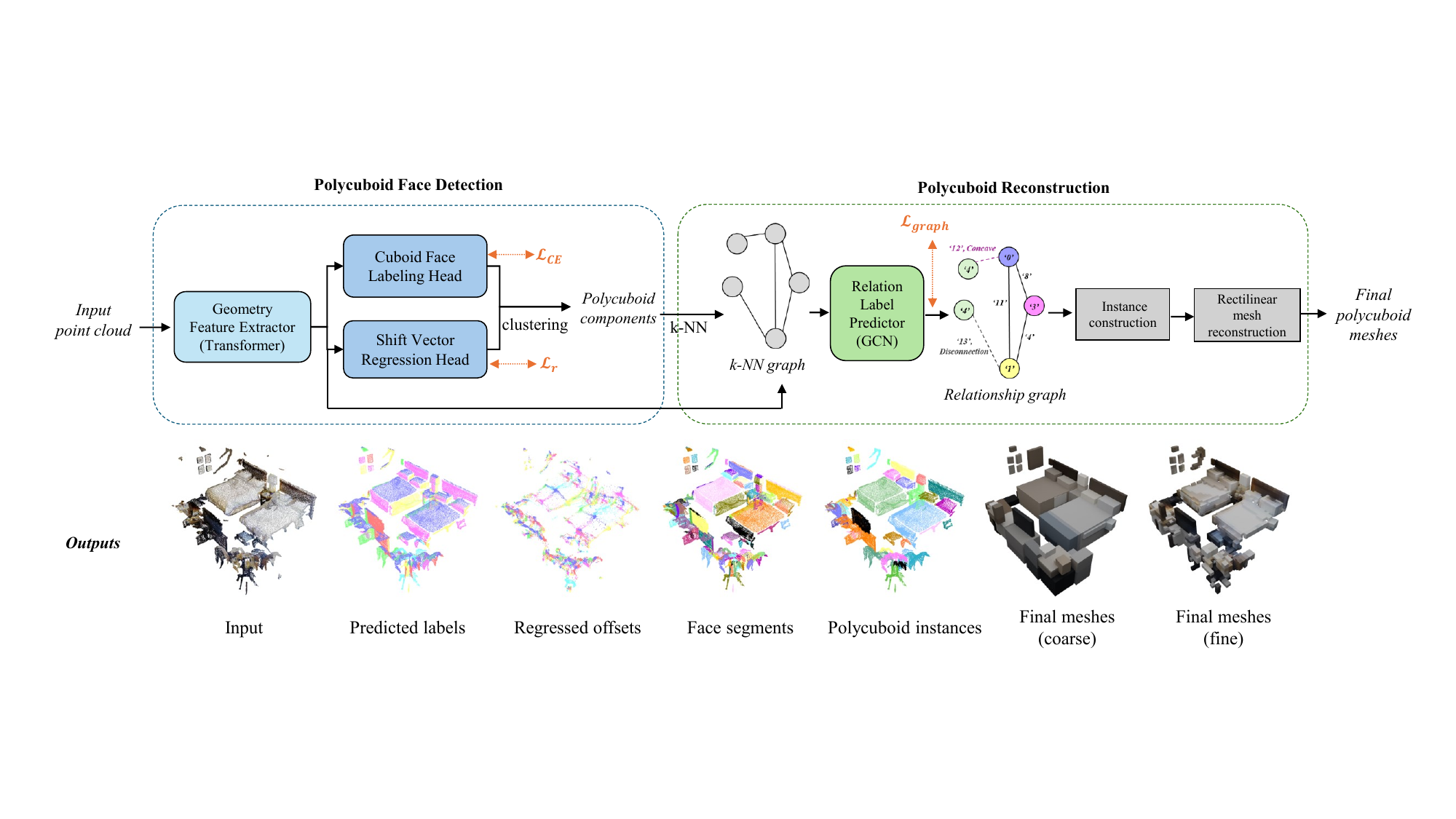}
\end{center}
   \caption{Overall process of our polycuboid fitting framework. 
   Initially, a transformer network estimates point-wise face labels and offsets that are used to detect face segments. Detected face segments are then aggregated into polycuboid instances using spatial relations predicted by a graph convolutional network. Finally, each polycuboid instance is reconstructed as a polycuboid mesh in a coarse or fine detail level.} 
\label{fig:overall}
\end{figure*}

\section{Related Work}
 
\paragraph*{Cuboid fitting} 
There have been studies that leverage cuboid shapes to represent indoor scenes or man-made objects \cite{shao2014imagining, mishima2018rgb, he2021manhattan, jiang2013linear, ramamonjisoa2022monteboxfinder}.
Some methods \cite{ramamonjisoa2022monteboxfinder, mishima2018rgb, jiang2013linear} detect orthogonal planar segments to establish a set of cuboids.
He et al. \cite{he2021manhattan} incorporates corners for cuboid fitting that are generated by intersecting lines and planes extracted from an image.
Tianjia et al. \cite{shao2014imagining} proposed a method that first fits cuboids to separated point clusters and then optimizes the cuboid parameters. 
However, most previous methods rely on heuristic plane fitting and generate undesirable cuboids like planar boxes, requiring post-processing to filter out invalid cuboids. 
Our approach fully explores important geometric constraints within a cuboid, such as adjacency relationships between cuboid faces. This enables robust generation of plausible polycuboids from a noisy point cloud without post-processing. 

\paragraph*{Primitive-based object abstraction}
Parsing 3D scanned objects into simple primitives, such as planes, cylinders, spheres, superquadrics, and cuboids, has been studied in the context of shape abstraction and reconstruction.
PolyFit~\cite{nan2017polyfit} introduced a framework for reconstructing polygonal surfaces from noisy points by selecting outer faces from a set of hypothesized faces. 
A few works \cite{li2019supervised, yan2021hpnet, sharma2020parsenet} have learned to fit various primitive types to a point cloud for representing an object. 
Recently, superquadrics have regained attention due to their ability to achieve low fitting errors~\cite{paschalidou2019superquadrics, liu2022robust, wu2022primitive}.
A cuboid has long been emphasized as a geometric primitive due to its significant benefit in representing manufactured objects  
~\cite{tulsiani2017learning, sun2019learning, yang2021unsupervised}.
Despite their impressive results, these methods assume 
a clean input geometry
and are limited to object-scale data.

\paragraph*{Primitive-based scene abstraction}
A scene typically contains various man-made objects that often exhibit piecewise smooth or planar surfaces. 
Given this characteristic, several studies have employed simple primitives, such as lines and planes, to approximate the underlying geometry of the target scene.
Zhou et al. \cite{zhou2019wirefrmae} reconstruct a 3D wire-frame from a single RGB image by leveraging 2D geometric cues such as junctions, lines, and vanishing points based on the Manhattan assumption. 
Cuboids have also been widely adopted as primitives for scene-scale abstraction \cite{shao2014imagining, mishima2018rgb, jiang2013linear,  ramamonjisoa2022monteboxfinder}. 
Especially, similar to ours, MonteBoxFinder \cite{ramamonjisoa2022monteboxfinder}  detects cuboids in a scene-scale point cloud.
However, previous cuboid fitting methods are constrained to representing only the convex regions, making it challenging to capture concave objects in a scene. In contrast, our polycuboid fitting method effectively represents both convex and concave objects, enhancing the overall representation capability.

 
\section{Deep Polycuboid Fitting}
\label{sec:keyidea} 
Our framework converts a point cloud of an indoor scene into a set of polycuboids for abstract scene representation in two key steps: polycuboid component detection (\Sec{polycube component detection}) and polycuboid reconstruction (\Sec{polycuboid reconstruction}).
In the polycuboid component detection stage, we first infer a cuboid face label and a shift vector for each 3D point in the input point cloud. Using this information, we cluster points to detect polycuboid face components via a density-based algorithm.
In the polycuboid reconstruction stage, we infer fourteen predefined spatial relationships within a polycuboid for the detected faces by using a graph convolutional network (GCN). 
Based on the inferred spatial relationships, connected faces are aggregated into polycuboid instances. Finally, each polycuboid is reconstructed as a set of cuboids using the label information from the aggregated faces. This process is illustrated in \Fig{overall}. 
 
\subsection{Polycuboid Component Detection } 
\label{sec:polycube component detection}
 
We adopt a bottom-up approach to detect polycuboid face components from an input point cloud. Our method infers a cuboid face label and a center shift vector for each 3D point, and this information is used to cluster points, where each cluster represents a face component.

\begin{figure}[t]  
\begin{center} 
     \begin{subfigure}[b]{0.2\columnwidth}
         \includegraphics[width=\textwidth]{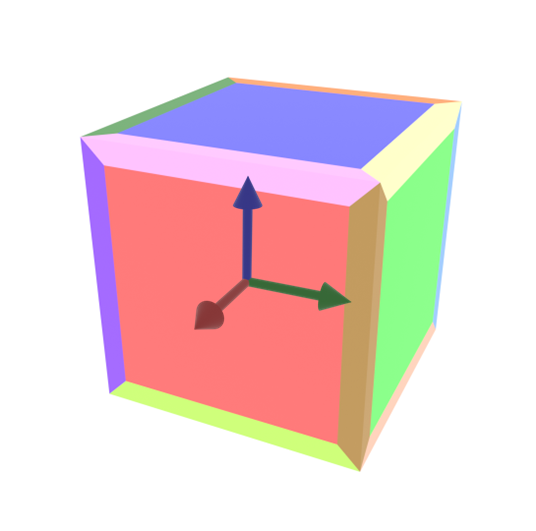}
          \caption{} 
     \end{subfigure}
     \begin{subfigure}[b]{0.18\columnwidth}
         \includegraphics[width=\textwidth]{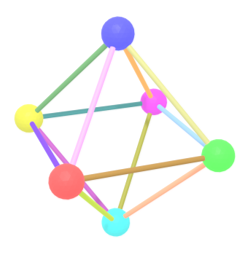}
          \caption{} 
     \end{subfigure}
     \begin{subfigure}[b]{0.2\columnwidth}
         \includegraphics[width=\textwidth]{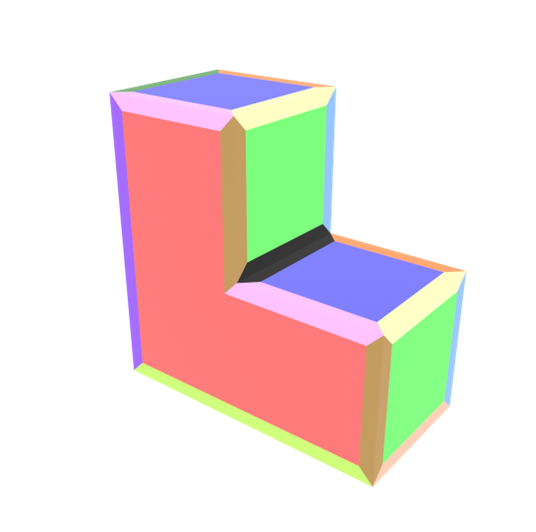}
          \caption{} 
     \end{subfigure}
        \caption{Face labels and their spatial relationships in a cuboid, and their extension to a polycuboid.
        (a) A cuboid with color-coded face and edge components. (b) Spatial relationship graph of cuboid components, whose nodes and edges correspond to faces and edges of a cuboid, respectively. (c) A polycuboid with an additional adjacency type of concave edge marked in black.}
        \label{fig:cube system} 
\end{center}
\end{figure}

\paragraph*{Point-wise cuboid face labeling}  
For cuboid face labeling, we formulate a multiclass classification problem where each class represents one of the six cuboid faces. 
As illustrated in \Fig{cube system}, any face of a polycuboid can be assigned one of six cuboid face labels (corresponding to the six directional axes: $\pm x, \pm y, \pm z$).
To implement the labeling task, we employ the Stratified Transformer~\cite{lai2022stratified} as our backbone network, which is especially designed for scene geometry understanding of large-scale point clouds.  

Given an input point cloud, the backbone network extracts point-wise features \textbf{F}. These features are then passed through a two-layer MLP head to predict probability vectors $\mathbf{s}\in\mathbb{R}^{6}$ for cuboid face labels. The final label for each point is determined by selecting the label with the highest probability. The backbone and head networks are trained using a cross-entropy loss defined as
\begin{equation}
\mathcal{L}_{CE}={\frac{1}{N}}\sum{CE(\mathbf{s}_i,\mathbf{s}_i^*)},
\end{equation}
where $N$ is the number of points and $\mathbf{s}^*$ is the ground-truth label.

\paragraph*{Point-wise shift vector regression} 
Assigning cuboid face labels alone is not sufficient for accurately detecting 
polycuboid faces.
3D points with the same cuboid face label but belonging to different cuboids should not be grouped into the same face.
Rather than relying on a heuristic distance threshold, we employ a neural network to regress shift vectors, enabling robust detection of distinct faces. 

Shift vector regression has been widely used for instance-level association in object detection and segmentation~\cite{qi2019deep, jiang2020pointgroup, Vu_2022_CVPR}. In our case, the shift vector $\mathbf{o}\in\mathbb{R}^{3}$ represents the displacement of each point to the center of its corresponding polycuboid face. 
Similar to face labeling, shift vector regression utilizes a shared backbone network and a two-layer MLP head. The network is trained using $\mathit{l}_2$ regression loss defined as 
\begin{equation}
\mathcal{L}_{r}={\frac{1}{N}}\sum{ \| \mathbf{o}_i - \mathbf{o}_i^* \|_2 },
\end{equation}
where  
$\mathbf{o}^*$ is the ground-truth shift vector.

\paragraph*{Polycuboid face detection}  
Using the estimated face labels and shift vectors, we detect polycuboid faces by grouping spatially adjacent points that share the same labels. 
For each set of points sharing the same cuboid face label, points are translated along their shift vectors, aligning them near the centers of the corresponding polycuboid faces. The shifted points are then clustered using the DBSCAN algorithm~\cite{ester1996density} to identify individual polycuboid faces. 
Small polycuboid faces below a predefined size threshold (5cm in our implementation) are filtered out to remove noise.

\subsection{Polycuboid Reconstruction}
\label{sec:polycuboid reconstruction} 
We reconstruct polycuboid instances by aggregating detected faces and identifying cuboids that are compatible with the face configuration.
We first infer 
spatial relationships among the detected faces through a Graph Convolutional Network (GCN)~\cite{wald2020learning} to group them into individual polycuboid instances.
Each polycuboid instance is then reconstructed as a rectilinear mesh by assembling cuboids based on the grouped faces.




\paragraph*{Polycuboid instance construction} 
\label{sec aggregation} 
To infer the spatial relationships among the detected faces,
we first analyze these relationships within a polycuboid.
Since a cuboid is the building block of a polycuboid, the relationships can be derived from those of a cuboid.
A cuboid consists of six faces, each maintaining a fixed adjacency relationship with its neighboring faces.
For example, in \Fig{cube system}a, the red face is always adjacent to the front side of the purple face. 
Likewise, a cuboid has twelve distinct spatial relationships among faces, which correspond to twelve cuboid edges.
These relationships can be represented as a graph (\Fig{cube system}b).

To represent spatial relationships among faces within a polycuboid, we 
incorporate two additional adjacency types. 
{\em Concave} adjacency refers to two faces that meet at an inward angle, forming a non-convex structure (\Fig{cube system}c).
{\em Disconnected} adjacency refers to two faces that are spatially separated and do not touch each other.
With these two additional types, the spatial relationship between any pair of detected faces can be categorized into one of fourteen types.

We infer the spatial relationships of detected faces using a graph representation.
In the graph, each node corresponds to a detected face, and edges are connected to the $k$-nearest neighbor ($k$-NN) faces based on Euclidean distances.
Given this graph as the input, we use a GCN to infer the final face label for each node and the adjacency type for each edge.
Note that the face labels determined in the previous polycuboid face detection step may be erroneous due to noise, partial scans, and rotational ambiguity. 
These errors can be corrected by gathering information from neighboring faces through a GCN.

For each node in the graph, we store 256 points randomly sampled from 3D points belonging to the corresponding detected face.
We also concatenate the features $F$ to the sampled points, which have been computed by the backbone network during the face labeling step.
Similarly, for each edge in the graph, we store the concatenation of points and their features from the two connected nodes. 
We then use two separate PointNet~\cite{qi2017pointnet} encoders to extract features for nodes and edges, respectively.
We use a similar approach to~\cite{wald2020learning} to train and infer a GCN using the graph.

Our GCN predicts the probability vectors for six possible face labels for node and fourteen possible adjacency types for edge, denoted by $\mathbf{n}\in\mathbb{R}^{6}$ and $\mathbf{e}\in\mathbb{R}^{14}$, respectively.
The PointNets and GCN are trained using a combination of two cross-entropy losses defined as
\begin{equation}
\mathcal{L}_{graph}={\frac{1}{N_{n}}}\sum{CE(\mathbf{n}_i,\mathbf{n}_i^*)} + {\frac{1}{N_{e}}}\sum{CE(\mathbf{e}_i,\mathbf{e}_i^*)},
\end{equation}
where $N_n$ is the number of nodes and $N_e$ is the number of edges, and $\mathbf{n}^*$ and $\mathbf{e}^*$ are the ground-truth labels of nodes and edges, respectively.
The trained GCN can robustly infer the final face labels and their spatial relationships to form polycuboid instances. 

Using the inference results of the GCN, we group detected faces into polycuboid instances.
For each edge in the graph, we verify whether its adjacency type and the face labels of the two connected nodes conform to the predefined cuboid label configuration in \Fig{cube system}b.
We then extract subgraphs that remain after removing edges with invalid configurations, obtaining valid polycuboid instances. 

\begin{figure}
\begin{center}
\includegraphics[width=1\linewidth, trim=0cm 4cm 4cm 0cm]{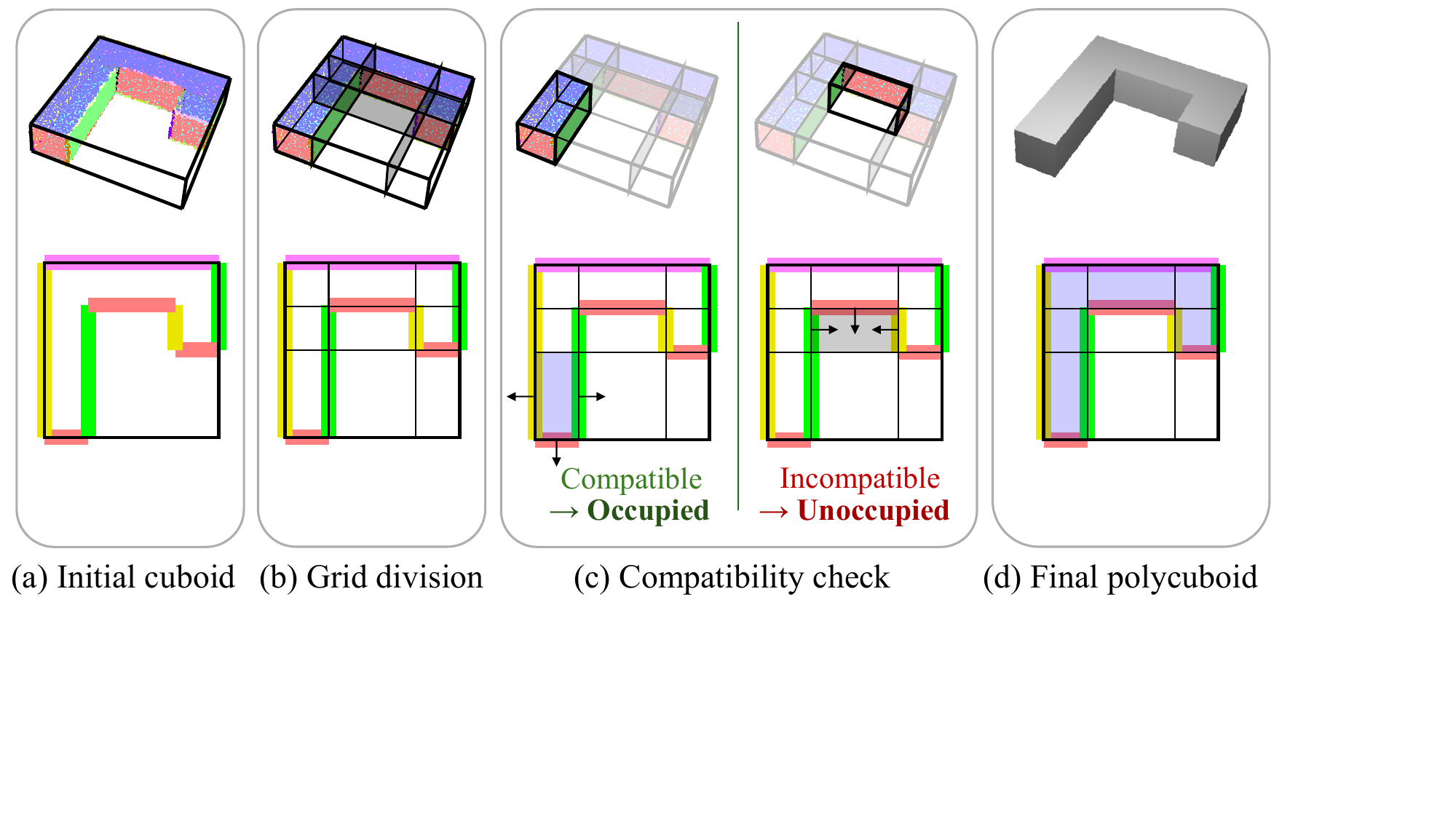}
\end{center}
   \caption{Illustration of our polycuboid reconstruction process. From a 3D non-uniform grid fitted to a polycuboid instance, we select valid 3D boxes that are inside the polycuboid based on the inferred face labels.}  
\label{fig:polyrecon}
\end{figure}
  
\paragraph*{Rectilinear mesh reconstruction}  
Finally, we reconstruct a compact polycuboid mesh for each polycuboid instance.
As a polycuboid instance may be slightly rotated relative to the global frame, we first align its local frame with the global frame before applying our reconstruction process. After reconstruction, we transform the mesh back to its original orientation.

To generate a polycuboid mesh, we fit a bounding box to the polycuboid instance and construct a 3D non-uniform grid by slicing the bounding box with planes fitted to the detected faces (\Fig{polyrecon}ab). 
A subset of 3D boxes in the grid is then selected to determine the shape of the polycuboid instance. 
We perform the box selection by checking the label configuration of detected faces surrounding each box.
Only the boxes with compatible configurations are classified as inside the polycuboid instance.

For checking the compatibility of face configuration,
we introduce a heuristic score function.
Boxes with positive scores are classified as parts of the polycuboid instance, while those with negative scores are excluded. The final polycuboid mesh is 
reconstructed 
from the outer surface of the selected 3D boxes.
Details on the score function are provided in the supplementary material.
While polycuboid meshes provide a compact abstraction of the scene, more details may be required for certain applications. Finer geometric details can be obtained by subdividing 
the 3D non-uniform grid and reapplying the reconstruction process. 
We set the subdivision interval to 0.1m in our experiments.
For visualization, each box is colored using the average color of the  
surrounding points of its faces.
%


\subsection{Indoor Scene Polycuboid Reconstruction} 

\begin{figure}
\begin{center}
\includegraphics[width=1\linewidth, trim=0cm 7cm 4cm 0cm]{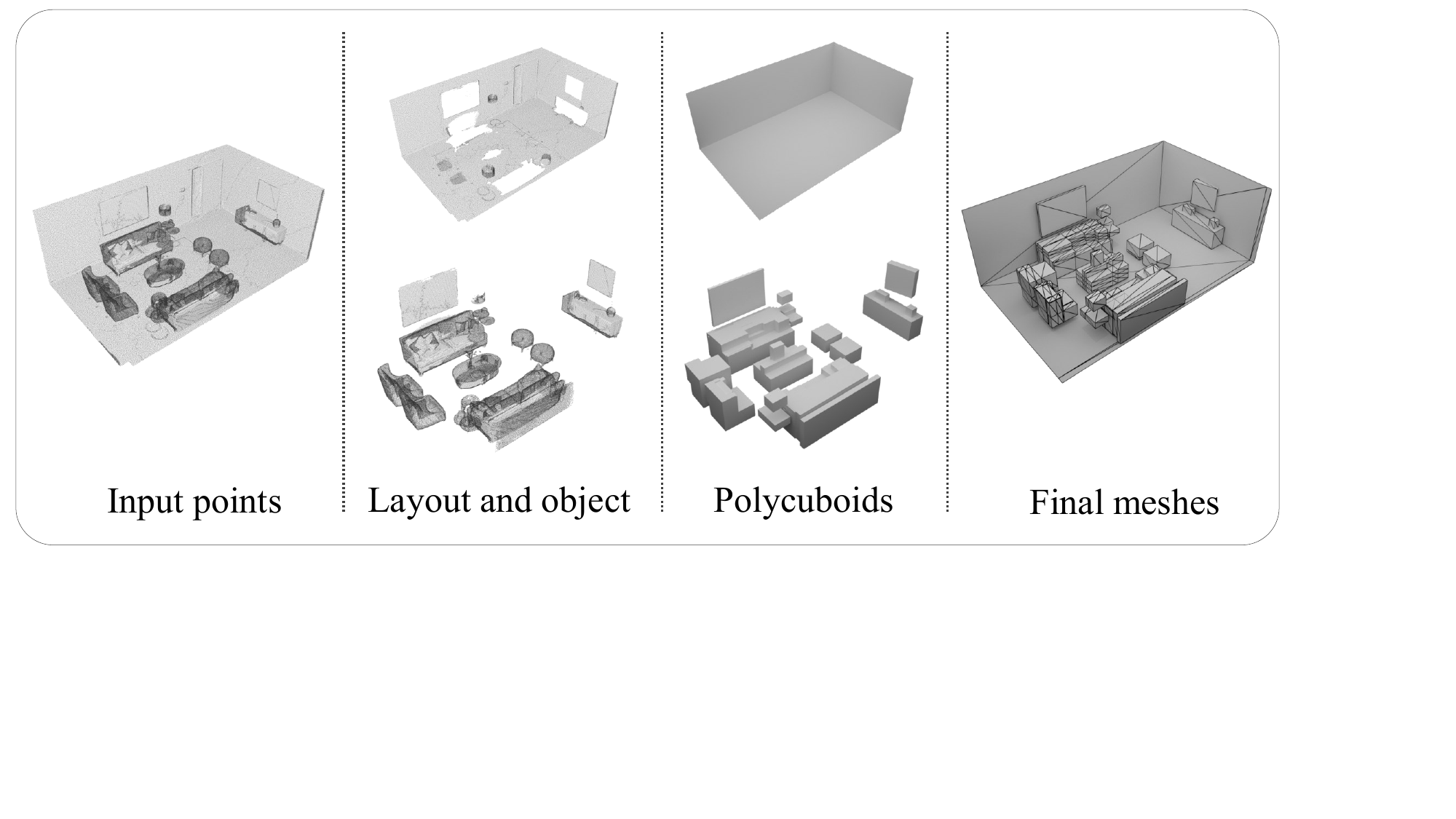}
\end{center}
   \caption{Illustration of our indoor scene reconstruction pipeline. The input point cloud is first separated into layout and object points. These two point sets are independently reconstructed into polycuboid meshes using our polycuboid fitting method, which are then merged to produce the final output.   
} 
\label{fig:indoor_recon}
\vspace*{-0.06cm}
\end{figure}

Indoor scenes typically consist of objects placed in the room layout, forming a nested structure.
To avoid potential ambiguity among object- and room-level structures,
we divide the scene into two parts: the room layout and the objects within it.  
To separate the room layout from the input point cloud, we employ an off-the-shelf point cloud segmentation method~\cite{lai2022stratified}. We then apply our pipeline to the room layout and object parts separately (\Fig{indoor_recon}).  


\section{Experiments}

\subsection{Datasets}
\label{sec:dataset}
We build a synthetic \textit{Polycuboid} dataset for training our networks and conduct qualitative and quantitative evaluations on three distinct real world datasets: \textit{ScanNet~\cite{dai2017scannet}, Replica~\cite{replica19arxiv}}, and \textit{iPhoneCapture.} 

\paragraph*{Polycuboid dataset}
We introduce a synthetic polycuboid dataset designed to replicate the 3D structures of indoor scenes and real-world scanning conditions. The dataset contains 1,800 training and 200 validation scenes, each composed of a mixture of polycuboid and cuboid meshes.
To provide diverse contextual information, we generate two types of configurations; 1) Polycuboids with random arrangements, 2) Polycuboids with contextual arrangements, derived from bounding boxes provided in the ScanNet dataset. We then sample 3D points at 1cm intervals on each face of the meshes.
To mitigate domain gaps caused by a scanning process, inspired by DoDA \cite{ding2022doda}, we simulate scanning noise by adding Gaussian noise and creating holes by randomly removing faces and their corresponding points. 
For the face component detection task, each point is assigned a label corresponding to one of six face labels and an offset vector pointing to its respective face center. For the graph prediction task, we reconstruct individual faces from the sampled points and build a k-nearest neighbors (k-NN) graph based on these faces. In this graph, each node is assigned one of six labels, while each edge is given one of fourteen possible labels. Details on data generation are provided in the supplementary material. 

\paragraph*{ScanNet} 
ScanNet dataset \cite{dai2017scannet} consists of over 1,500 3D indoor scene scans, providing annotated bounding boxes for objects in the scene.  
We use the validation set for our evaluation. Since points corresponding to layout categories are too incomplete for reconstruction, we only use points classified as object categories for evaluation.  

\paragraph*{Replica} 
Replica dataset \cite{replica19arxiv} features highly photo-realistic digitization of real indoor scenes, represented with clean and dense meshes. Among 18 different scenes, excluding FRL apartment scenes, we use a total of 9 scenes consisting of 5 office rooms, a hotel room, and 3 rooms of apartments for evaluation.  

\begin{figure}
\begin{center}
\includegraphics[width=\linewidth,  trim=1cm 9cm 8cm 0cm]{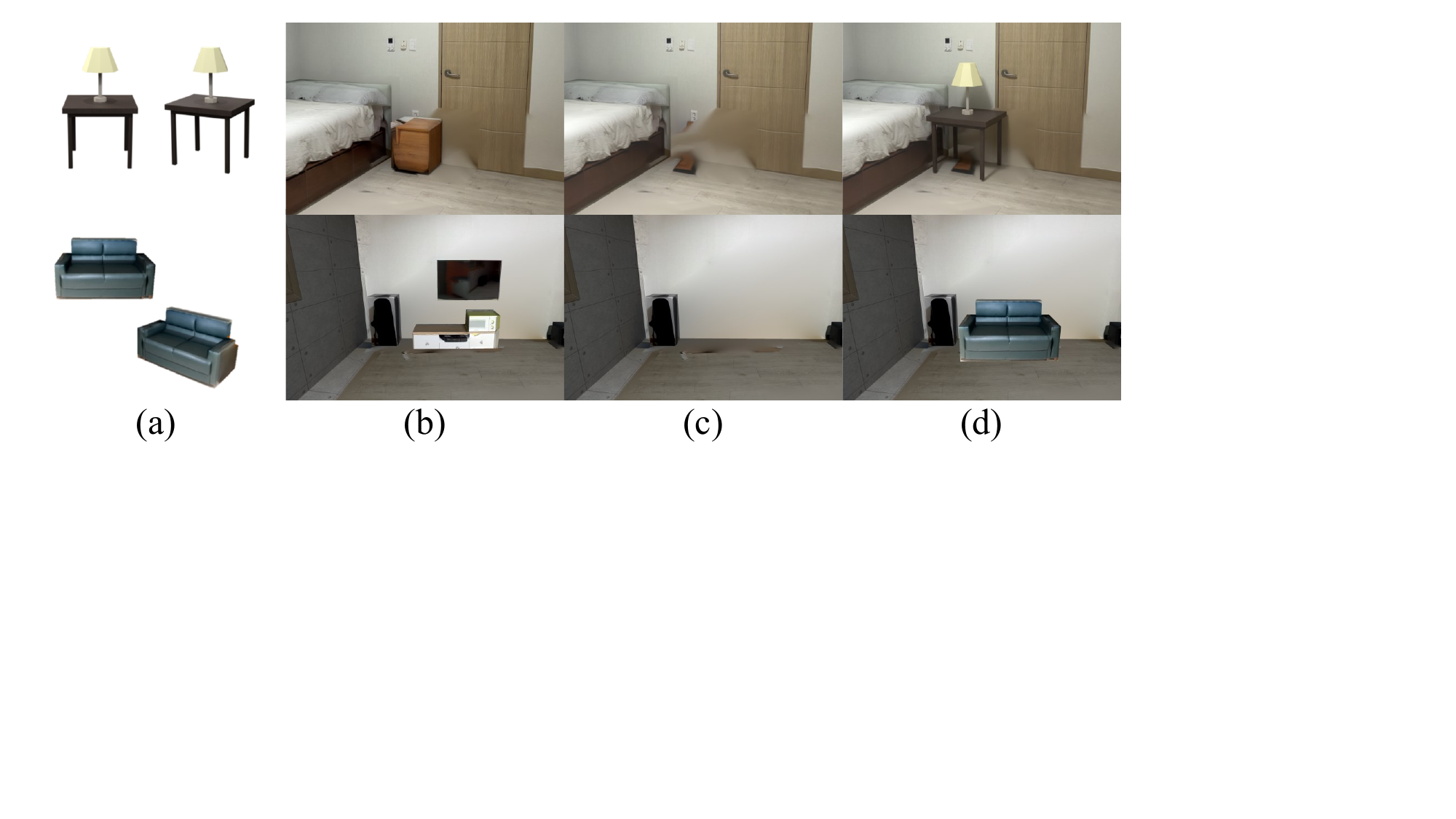}
\end{center}
  \caption{Room editing application for object removal and rearrangement. (a) New objects to be placed into the scene, bedside table and sofa. (b) Original scene reconstructed with polycuboids. (c) Scene after object removal. (d) Scene with newly added object. 
  }
\label{fig:app_editing}
\end{figure}

\paragraph*{iPhoneCapture}
We demonstrate the effectiveness of our polycuboid reconstruction on scenes captured by a mobile phone. We used Polycam \cite{polycam}, a commercial 3D scanning app, to capture two common real-world room types, a living room and a bedroom, with an iPhone Pro 12. 
We present two demos to showcase the potential of our method in practical applications, such as virtual room tour and room editing (\Figs{quali_iphone} and \ref{fig:app_editing}).  

\subsection{Implementation Details}   
We train the transformer~\cite{lai2022stratified} and GCN~\cite{wald2020learning} using the Polycuboid dataset. The transformer is trained with the AdamW~\cite{loshchilov2017decoupled} optimizer at a learning rate of 0.006, a weight decay of 0.05, and a batch size of 4 over 100 epochs, using a voxel size of 0.02m. The GCN is trained with the Adam optimizer at an initial learning rate of 0.0001, adjusted at epochs 30 and 60, with a weight decay of 0.5. GCN training runs for 80 epochs with a batch size of 8, using a $k$-NN graph with $k$ set to 5. All training is conducted on a single NVIDIA GeForce RTX 4090 GPU. 

\subsection{Results and analysis} 
\label{sec:evaluation}

\paragraph*{Baseline}
Our framework is designed for scene-scale 3D abstraction of an input point cloud, in contrast to most existing studies that focus on object-scale abstraction. We compare our method with MonteBoxFinder (MBF)~\cite{ramamonjisoa2022monteboxfinder} as a scene-scale abstraction work, which employs cuboid fitting for input point clouds representing indoor scenes.

\begin{figure*}[t]
\begin{center}
  \includegraphics[width=0.95\linewidth,trim=1cm 1cm 9cm 0cm]{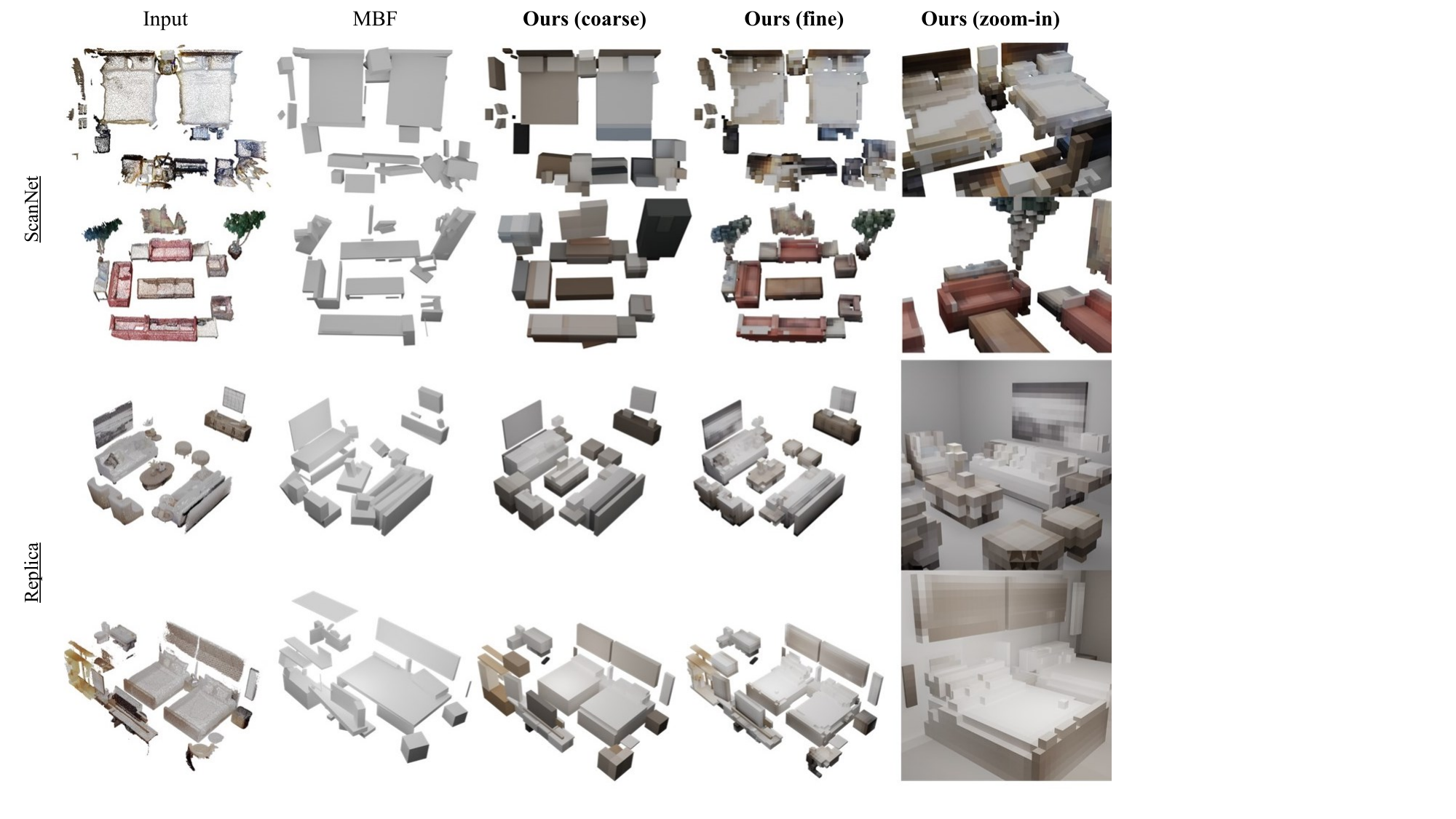}
\end{center}
\caption{Qualitative comparison on ScanNet and Replica datasets. Columns from left to right show the input point clouds, results from MBF\cite{ramamonjisoa2022monteboxfinder}, results from our method in coarse and fine detail levels, and zoomed-in views of the fine level results.
}
\label{fig:quali}
\end{figure*}

\begin{figure*}[t]
\begin{center}
\includegraphics[width=\linewidth, trim=1cm 9cm 1cm 0cm]{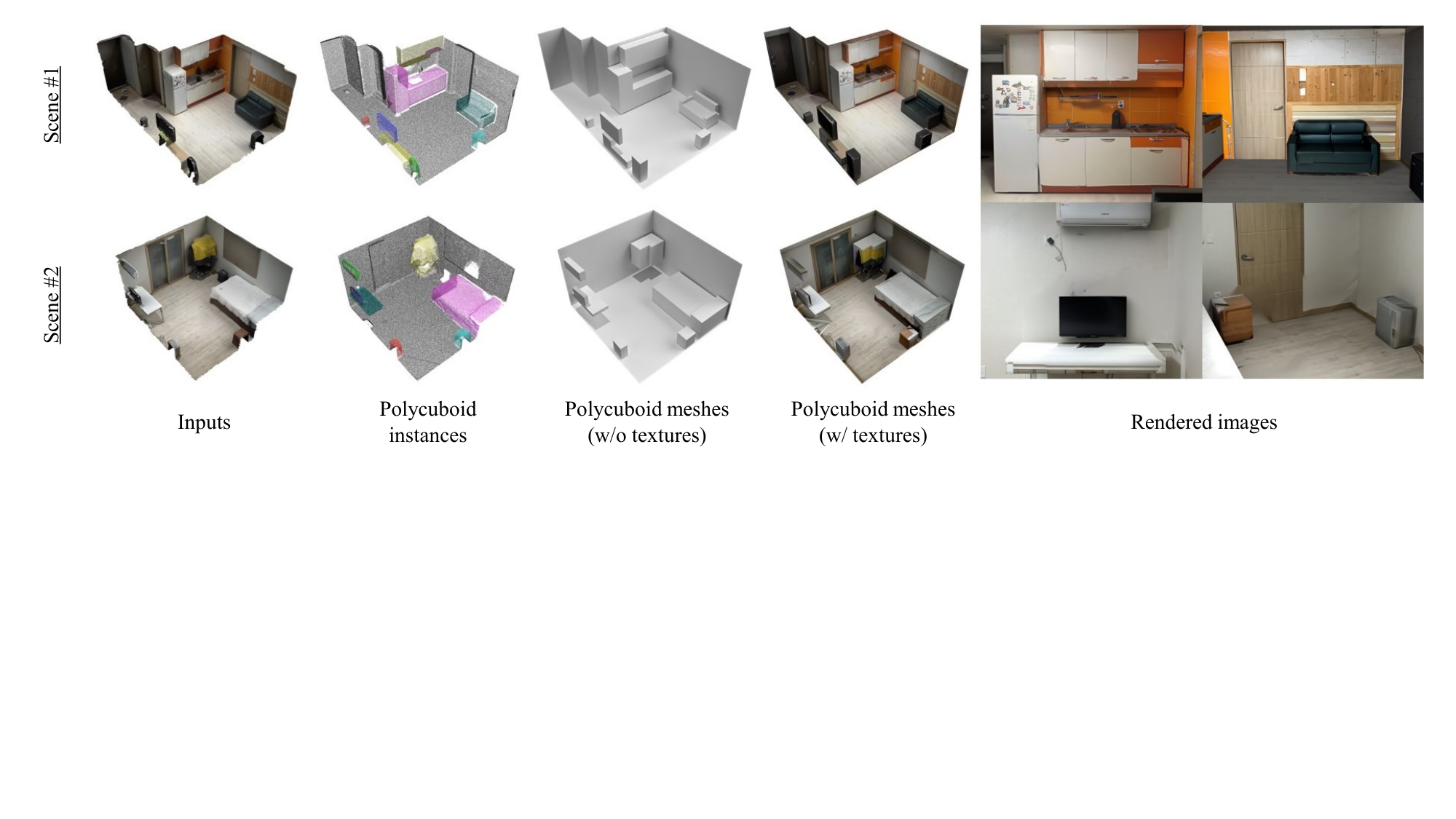}  
\end{center}
   \caption{Qualitative evaluations on room data captured using an iPhone. Columns from left to right show the input point clouds, point sets covered by detected polycuboid instances encoded with different colors, reconstructed polycuboids without and with textures, and rendering results.
}
\label{fig:quali_iphone}
\end{figure*}

\paragraph*{Reconstruction fidelity}  
We evaluate the reconstruction fidelity of our method by   
utilizing the Chamfer Distance (CD) metric which quantifies the discrepancy between the input point cloud and the reconstructed polycuboids. 
Even though real world data includes severe holes (e.g., walls and furniture with large hidden parts) and moderate noise, our method effectively handles these challenges and produce polycuboids faithfully recovering the underlying geometry of input point clouds, as illustrated in \Figs{quali} and \ref{fig:quali_iphone}. 
In contrast, MBF, which relies on a heuristic cuboid fitting approach, often produces thin planar boxes that fail to accurately represent objects, as shown in \Fig{quali_replica_zoomin}.

Our approach achieves better quantitative results for non-layout objects, while MBF slightly outperforms in metrics for layout parts. This is likely due to structured noise around transparent materials, like windows, in layout categories, causing our method to generate slightly larger polycuboids. Despite this, our method produces visually plausible results for layout points, as shown in \Figs{indoor_recon} and \ref{fig:quali_iphone}. 

Regarding the level of polycuboid details, we provide abstraction results at a higher resolution in the last two columns of \Fig{quali}. This resolution can be adjusted by the user for different purposes, depending on the application.
More qualitative and quantitative results can be found in the supplementary material.

\begin{table}
\caption{Quantitative evaluation of reconstruction fidelity. We measure Chamfer Distance (CD) on ScanNet and Replica datasets.}
\label{table:comp_quanti_cd}
\begin{center}
\begin{tabular}{lccc}
\toprule
        & \multicolumn{1}{c}{ScanNet} & \multicolumn{2}{c}{Replica}    \\
\cmidrule(rl){2-4}  
  &  non-layout  &   non-layout &   layout  \\
\hline
MBF\cite{ramamonjisoa2022monteboxfinder} & 0.066 & 0.078 &  \textbf{0.047} \\
\textbf{Ours} &  \textbf{0.040} & \textbf{0.044}  & 0.081 \\
\hline
\toprule
\end{tabular}
\end{center}
\end{table}

\begin{table}
\begin{center}
\caption{Quantitative evaluation of object-level representation. We measure precision and recall scores on ScanNet validation set.}
\label{table:scannet}
\begin{tabular}{lcccc}
\toprule
        & \multicolumn{2}{c}{IoU 25} & \multicolumn{2}{c}{IoU 50}    \\
\hline
Method &  Recall & Precision   &  Recall & Precision  \\
\hline 
MBF\cite{ramamonjisoa2022monteboxfinder} & 0.41 & 0.11  & 0.11 & 0.03  \\
\textbf{Ours} &   \textbf{0.64} &  \textbf{0.15} & \textbf{0.43}  &  \textbf{0.10}  \\
\hline
\toprule
\end{tabular}
\end{center}
\end{table}

\paragraph*{Object-level representation}  
We assess how effectively our fitted polycuboids approximate scenes at the object level in the ScanNet dataset.
We compute precision and recall scores on detected polycuboids using ground truth bounding boxes at two Intersection over Union (IoU) thresholds, 25 and 50, respectively.  
As shown in Table. \ref{table:scannet}, 
our method achieves higher scores than MBF, demonstrating that our approach more accurately decomposes the scene, with each polycuboid more likely representing a single object.
As MBF tends to over-decompose the scene, its recall score drops significantly as the IoU threshold increases. In contrast, our method demonstrates robust performance, maintaining higher recall scores even at a stricter IoU threshold.

\begin{figure}[t]  
\begin{center}
  \includegraphics[width=1.05\linewidth, trim=1cm 9cm 5cm 0cm, clip]{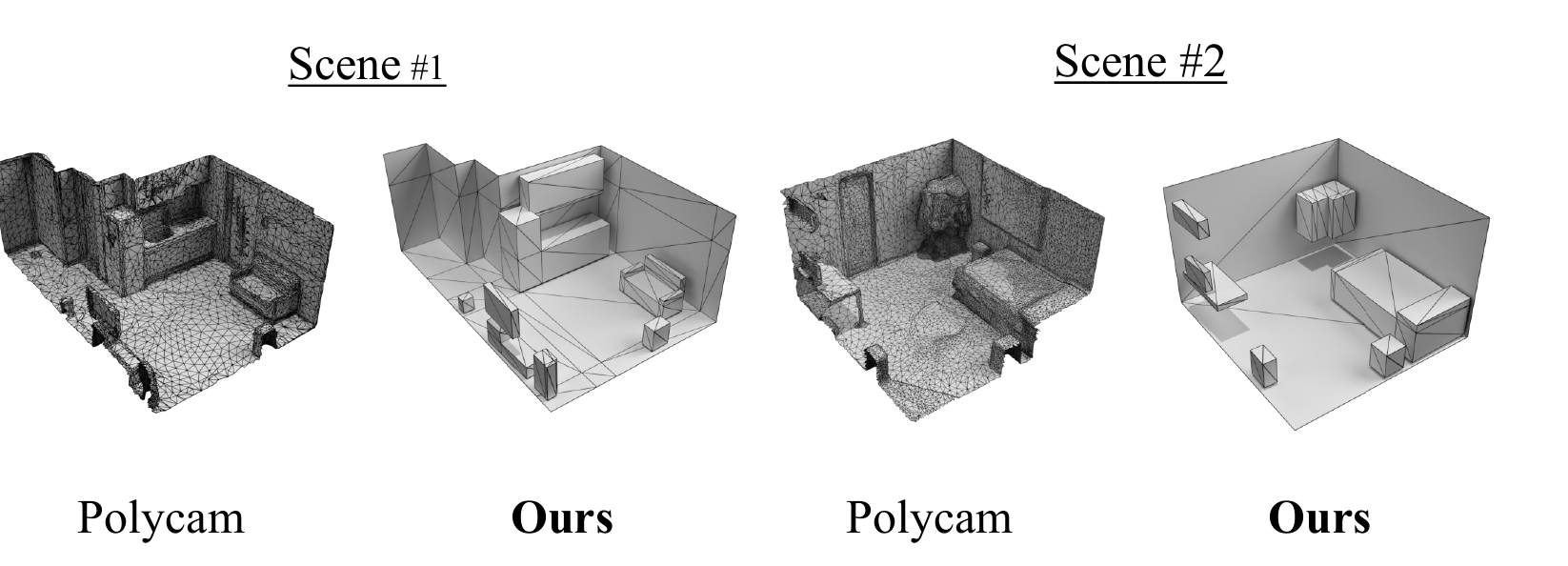}
\end{center}
\caption{Comparison of reconstruction efficiency with a commercial application~\cite{polycam} on data captured using an iPhone. Our method provides the benefit of a lightweight representation.} 
\label{fig:lightweight}
\end{figure}

\paragraph*{Potential applications}  

When combined with off-the-shelf texture mapping features provided by Blender, polycuboid geometry allows a visually plausible appearance of the scene to be obtained with a small number of polygons. 
To showcase the potential of our method in practical scenarios, we present two applications. 
\begin{itemize}   
\item{\textbf{Virtual room tour}} We can navigate the reconstructed rooms for seamless virtual exploration experience, as shown in \Fig{quali_iphone}, where polycuboid-based reconstruction provides complete and lightweight geometry. With texture mapping, realistic rendering can be achieved, closely resembling the original scanned data.
\item{\textbf{Room editing}} 
Each polycuboid represents a scene object such as a sofa or a TV shelf. The scene can be easily edited by manipulating the polycuboids, which would be useful for interior design and AR/VR applications.
\end{itemize}

\begin{figure}[t]  
\begin{center}
  \includegraphics[width=\linewidth,  trim=1cm 4cm 1cm 0cm]{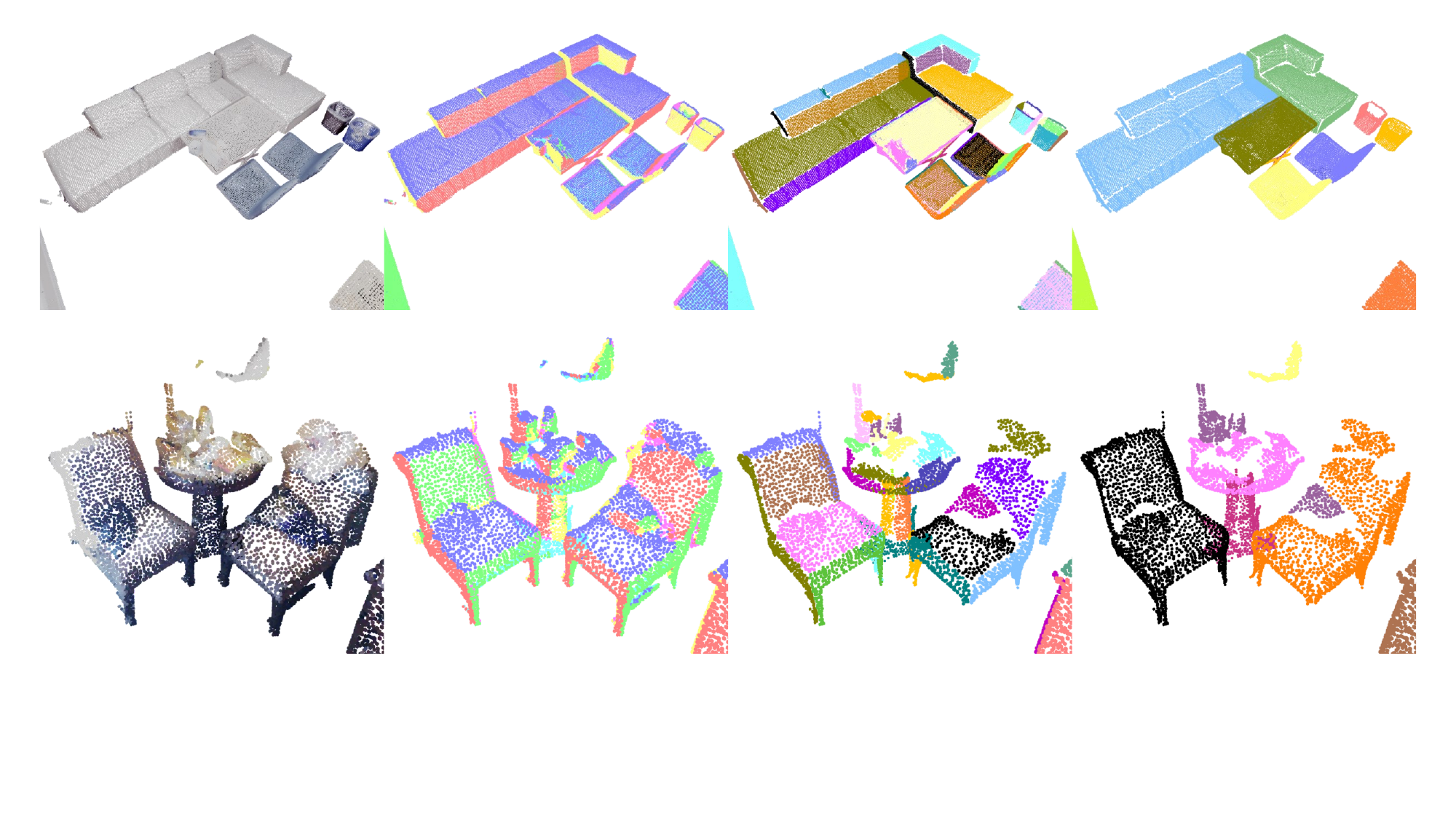}
\end{center}
\caption{Visualization of the intermediate result of each module in our framework on ScanNet data. The first column represents the input point clouds. In the remaining three columns, color encodings visualize estimated face labels, detected faces, and detected polycuboid instances, respectively.  
}
\label{fig:ablation}
\end{figure}

\subsection{Discussion}   
\label{sec discussions}

\paragraph*{Generalization ability on arbitrary shapes}   
Despite being trained solely on the synthetic Polycuboid dataset, our framework generalizes effectively to real-world datasets with various shapes and scanning conditions.
We visualize the intermediate result from each module, including face detection and polycuboid instance construction.
As shown in Fig. \ref{fig:ablation}, the transformer network consistently assigns cuboid face labels to input points, demonstrating robustness in detecting object faces even with complex and noisy geometries. 
In addition, our GCN provides concrete inference results needed for robust reconstruction of polycuboid instances. 

\paragraph*{Evaluation for lightweight representation}  
In \Fig{lightweight}, for scene \#1, Polycam~\cite{polycam} generates 15,011 vertices and 25,688 faces, while our polycuboid meshes have just 182 vertices and 328 faces. Similarly, for scene \#2, Polycam~\cite{polycam} produces 27,802 vertices and 48,196 faces, compared to our 94 vertices and 156 faces. Our method uses over 90 times fewer vertices and 60 times fewer faces, producing exceptionally lightweight yet complete meshes, compared to the already simplified outputs of Polycam~\cite{polycam}. This reduction in mesh topology would significantly improve computational efficiency for downstream applications such as VR and AR.

\paragraph*{Limitation}  
Due to the rectilinear nature of polycuboids, geometric discrepancies arise when abstracting objects with highly curved surfaces, which can 
potentially degrade visual fidelity.
A promising future work is to detect regions with high geometric errors and refine the polycuboids in those regions to enhance representation accuracy.
Also, our framework currently relies solely on geometric cues for polycuboid fitting. Incorporating semantic cues could further improve object-level abstraction and enhance reconstruction quality.

\section{Conclusion}  

In this paper, we proposed a framework for fitting polycuboids to indoor scenes based on geometric properties and configurations, without requiring high-level understanding of object structures.
In our framework, cuboid face labels and their spatial relationships, inferred from a transformer and a GCN, facilitate the reconstruction of polycuboid instances,
enabling a lightweight and editable representation of indoor scenes.
We demonstrate that our framework generalizes well to real-world datasets and has strong potential for practical applications.
We believe that leveraging primitive geometries, such as polycuboids, can benefit a wide range of scene representation and editing tasks.
\section*{Acknowledgements}
This work was supported by
NRF grants (RS-2023-00280400, RS-2024-00451947) 
and IITP grants (ICT Research Center, RS-2024-00437866; 
AI Innovation Hub, RS-2021-II212068; 
AI Graduate School Program, RS-2019-II191906) 
funded by Korea government (MSIT), 
and POSCO DX Company Ltd. (23DO1109)    .

\clearpage
{
    \small
    \bibliographystyle{ieeenat_fullname}
    \bibliography{main}

\begin{thebibliography}{35}
\providecommand{\natexlab}[1]{#1}
\providecommand{\url}[1]{\texttt{#1}}
\expandafter\ifx\csname urlstyle\endcsname\relax
  \providecommand{\doi}[1]{doi: #1}\else
  \providecommand{\doi}{doi: \begingroup \urlstyle{rm}\Url}\fi

\bibitem[Chekhlov et~al.(2007)Chekhlov, Gee, Calway, and
  Mayol-Cuevas]{chekhlov2007ninja}
Denis Chekhlov, Andrew~P Gee, Andrew Calway, and Walterio Mayol-Cuevas.
\newblock Ninja on a plane: Automatic discovery of physical planes for
  augmented reality using visual slam.
\newblock In \emph{2007 6th IEEE and ACM International Symposium on Mixed and
  Augmented Reality}, pages 153--156. IEEE, 2007.

\bibitem[Dai et~al.(2017)Dai, Chang, Savva, Halber, Funkhouser, and
  Nie{\ss}ner]{dai2017scannet}
Angela Dai, Angel~X Chang, Manolis Savva, Maciej Halber, Thomas Funkhouser, and
  Matthias Nie{\ss}ner.
\newblock Scannet: Richly-annotated {3D} reconstructions of indoor scenes.
\newblock In \emph{Proceedings of the IEEE conference on computer vision and
  pattern recognition}, pages 5828--5839, 2017.

\bibitem[Ding et~al.(2022)Ding, Yang, Jiang, and Qi]{ding2022doda}
Runyu Ding, Jihan Yang, Li Jiang, and Xiaojuan Qi.
\newblock Doda: Data-oriented sim-to-real domain adaptation for {3D} semantic
  segmentation.
\newblock In \emph{Computer Vision--ECCV 2022: 17th European Conference, Tel
  Aviv, Israel, October 23--27, 2022, Proceedings, Part XXVII}, pages 284--303.
  Springer, 2022.

\bibitem[Ester et~al.(1996)Ester, Kriegel, Sander, and Xu]{ester1996density}
Martin Ester, Hans-Peter Kriegel, J\"{o}rg Sander, and Xiaowei Xu.
\newblock A density-based algorithm for discovering clusters in large spatial
  databases with noise.
\newblock In \emph{Proceedings of the Second International Conference on
  Knowledge Discovery and Data Mining}, page 226–231. AAAI Press, 1996.

\bibitem[He et~al.(2021)He, Wang, and Cheng]{he2021manhattan}
Zhenbang He, Yunhai Wang, and Zhanglin Cheng.
\newblock Manhattan-world urban building reconstruction by fitting cubes.
\newblock In \emph{Computer Graphics Forum}, pages 289--300. Wiley Online
  Library, 2021.

\bibitem[Huang et~al.(2017)Huang, Dai, Guibas, and
  Nie{\ss}ner]{huang20173dlite}
Jingwei Huang, Angela Dai, Leonidas~J Guibas, and Matthias Nie{\ss}ner.
\newblock {3D}lite: towards commodity {3D} scanning for content creation.
\newblock \emph{ACM Trans. Graph.}, 36\penalty0 (6):\penalty0 203--1, 2017.

\bibitem[Jiang and Xiao(2013)]{jiang2013linear}
Hao Jiang and Jianxiong Xiao.
\newblock A linear approach to matching cuboids in rgbd images.
\newblock In \emph{Proceedings of the IEEE conference on computer vision and
  pattern recognition}, pages 2171--2178, 2013.

\bibitem[Jiang et~al.(2020)Jiang, Zhao, Shi, Liu, Fu, and
  Jia]{jiang2020pointgroup}
Li Jiang, Hengshuang Zhao, Shaoshuai Shi, Shu Liu, Chi-Wing Fu, and Jiaya Jia.
\newblock Pointgroup: Dual-set point grouping for {3D} instance segmentation.
\newblock In \emph{Proceedings of the IEEE/CVF conference on computer vision
  and Pattern recognition}, pages 4867--4876, 2020.

\bibitem[Kim and Woo(2022)]{kim2022integrating}
Yunho Kim and Honguk Woo.
\newblock Integrating a deep learning-based plane detector in mobile ar systems
  for improvement of plane detection.
\newblock In \emph{Proceedings of the 8th International Conference on Computing
  and Artificial Intelligence}, pages 597--602, 2022.

\bibitem[Lai et~al.(2022)Lai, Liu, Jiang, Wang, Zhao, Liu, Qi, and
  Jia]{lai2022stratified}
Xin Lai, Jianhui Liu, Li Jiang, Liwei Wang, Hengshuang Zhao, Shu Liu, Xiaojuan
  Qi, and Jiaya Jia.
\newblock Stratified transformer for {3D} point cloud segmentation.
\newblock In \emph{Proceedings of the IEEE/CVF Conference on Computer Vision
  and Pattern Recognition}, pages 8500--8509, 2022.

\bibitem[Li et~al.(2019)Li, Sung, Dubrovina, Yi, and Guibas]{li2019supervised}
Lingxiao Li, Minhyuk Sung, Anastasia Dubrovina, Li Yi, and Leonidas~J Guibas.
\newblock Supervised fitting of geometric primitives to {3D} point clouds.
\newblock In \emph{Proceedings of the IEEE/CVF Conference on Computer Vision
  and Pattern Recognition}, pages 2652--2660, 2019.

\bibitem[Lin et~al.(2013)Lin, Fidler, and Urtasun]{lin2013holistic}
Dahua Lin, Sanja Fidler, and Raquel Urtasun.
\newblock Holistic scene understanding for {3D} object detection with rgbd
  cameras.
\newblock In \emph{Proceedings of the IEEE international conference on computer
  vision}, pages 1417--1424, 2013.

\bibitem[Liu et~al.(2022)Liu, Wu, Ruan, and Chirikjian]{liu2022robust}
Weixiao Liu, Yuwei Wu, Sipu Ruan, and Gregory~S Chirikjian.
\newblock Robust and accurate superquadric recovery: A probabilistic approach.
\newblock In \emph{Proceedings of the IEEE/CVF Conference on Computer Vision
  and Pattern Recognition}, pages 2676--2685, 2022.

\bibitem[Loshchilov(2017)]{loshchilov2017decoupled}
I Loshchilov.
\newblock Decoupled weight decay regularization.
\newblock \emph{arXiv preprint arXiv:1711.05101}, 2017.

\bibitem[Mishima et~al.(2018)Mishima, Uchiyama, Thomas, Taniguchi, Roberto,
  Teichrieb, et~al.]{mishima2018rgb}
Masashi Mishima, Hideaki Uchiyama, Diego Thomas, Rin-ichiro Taniguchi, Rafael
  Roberto, Veronica Teichrieb, et~al.
\newblock {RGB-D} slam based incremental cuboid modeling.
\newblock In \emph{Proceedings of the European Conference on Computer Vision
  (ECCV) Workshops}, pages 0--0, 2018.

\bibitem[Nan and Wonka(2017)]{nan2017polyfit}
Liangliang Nan and Peter Wonka.
\newblock Polyfit: Polygonal surface reconstruction from point clouds.
\newblock In \emph{Proceedings of the IEEE International Conference on Computer
  Vision}, pages 2353--2361, 2017.

\bibitem[Paschalidou et~al.(2019)Paschalidou, Ulusoy, and
  Geiger]{paschalidou2019superquadrics}
Despoina Paschalidou, Ali~Osman Ulusoy, and Andreas Geiger.
\newblock Superquadrics revisited: Learning {3D} shape parsing beyond cuboids.
\newblock In \emph{Proceedings of the IEEE/CVF Conference on Computer Vision
  and Pattern Recognition}, pages 10344--10353, 2019.

\bibitem[Polycam(2022)]{polycam}
Polycam.
\newblock Polycam, 2022.

\bibitem[Qi et~al.(2017)Qi, Su, Mo, and Guibas]{qi2017pointnet}
Charles~R Qi, Hao Su, Kaichun Mo, and Leonidas~J Guibas.
\newblock Pointnet: Deep learning on point sets for {3D} classification and
  segmentation.
\newblock In \emph{Proceedings of the IEEE conference on computer vision and
  pattern recognition}, pages 652--660, 2017.

\bibitem[Qi et~al.(2019)Qi, Litany, He, and Guibas]{qi2019deep}
Charles~R Qi, Or Litany, Kaiming He, and Leonidas~J Guibas.
\newblock Deep hough voting for {3D} object detection in point clouds.
\newblock In \emph{proceedings of the IEEE/CVF International Conference on
  Computer Vision}, pages 9277--9286, 2019.

\bibitem[Ramamonjisoa et~al.(2022)Ramamonjisoa, Stekovic, and
  Lepetit]{ramamonjisoa2022monteboxfinder}
Micha{\"e}l Ramamonjisoa, Sinisa Stekovic, and Vincent Lepetit.
\newblock Monteboxfinder: Detecting and filtering primitives to fit a noisy
  point cloud.
\newblock In \emph{Computer Vision--ECCV 2022: 17th European Conference, Tel
  Aviv, Israel, October 23--27, 2022, Proceedings, Part XXVIII}, pages
  161--177. Springer, 2022.

\bibitem[Ruan et~al.(2022{\natexlab{a}})Ruan, Poblete, Wu, Ma, and
  Chirikjian]{ruan2022efficient}
Sipu Ruan, Karen~L Poblete, Hongtao Wu, Qianli Ma, and Gregory~S Chirikjian.
\newblock Efficient path planning in narrow passages for robots with
  ellipsoidal components.
\newblock \emph{IEEE Transactions on Robotics}, 39\penalty0 (1):\penalty0
  110--127, 2022{\natexlab{a}}.

\bibitem[Ruan et~al.(2022{\natexlab{b}})Ruan, Wang, and
  Chirikjian]{ruan2022collision}
Sipu Ruan, Xiaoli Wang, and Gregory~S Chirikjian.
\newblock Collision detection for unions of convex bodies with smooth
  boundaries using closed-form contact space parameterization.
\newblock \emph{IEEE Robotics and Automation Letters}, 7\penalty0 (4):\penalty0
  9485--9492, 2022{\natexlab{b}}.

\bibitem[Shao et~al.(2014)Shao, Monszpart, Zheng, Koo, Xu, Zhou, and
  Mitra]{shao2014imagining}
Tianjia Shao, Aron Monszpart, Youyi Zheng, Bongjin Koo, Weiwei Xu, Kun Zhou,
  and Niloy~J Mitra.
\newblock Imagining the unseen: Stability-based cuboid arrangements for scene
  understanding.
\newblock \emph{ACM Transactions on Graphics}, 33\penalty0 (6), 2014.

\bibitem[Sharma et~al.(2020)Sharma, Liu, Maji, Kalogerakis, Chaudhuri, and
  M{\v{e}}ch]{sharma2020parsenet}
Gopal Sharma, Difan Liu, Subhransu Maji, Evangelos Kalogerakis, Siddhartha
  Chaudhuri, and Radom{\'\i}r M{\v{e}}ch.
\newblock Parsenet: A parametric surface fitting network for {3D} point clouds.
\newblock In \emph{Computer Vision--ECCV 2020: 16th European Conference,
  Glasgow, UK, August 23--28, 2020, Proceedings, Part VII 16}, pages 261--276.
  Springer, 2020.

\bibitem[Straub et~al.(2019)Straub, Whelan, Ma, Chen, Wijmans, Green, Engel,
  Mur-Artal, Ren, Verma, Clarkson, Yan, Budge, Yan, Pan, Yon, Zou, Leon,
  Carter, Briales, Gillingham, Mueggler, Pesqueira, Savva, Batra, Strasdat,
  Nardi, Goesele, Lovegrove, and Newcombe]{replica19arxiv}
Julian Straub, Thomas Whelan, Lingni Ma, Yufan Chen, Erik Wijmans, Simon Green,
  Jakob~J. Engel, Raul Mur-Artal, Carl Ren, Shobhit Verma, Anton Clarkson,
  Mingfei Yan, Brian Budge, Yajie Yan, Xiaqing Pan, June Yon, Yuyang Zou,
  Kimberly Leon, Nigel Carter, Jesus Briales, Tyler Gillingham, Elias Mueggler,
  Luis Pesqueira, Manolis Savva, Dhruv Batra, Hauke~M. Strasdat, Renzo~De
  Nardi, Michael Goesele, Steven Lovegrove, and Richard Newcombe.
\newblock The {R}eplica dataset: A digital replica of indoor spaces.
\newblock \emph{arXiv preprint arXiv:1906.05797}, 2019.

\bibitem[Sun et~al.(2019)Sun, Zou, Tong, and Liu]{sun2019learning}
Chun-Yu Sun, Qian-Fang Zou, Xin Tong, and Yang Liu.
\newblock Learning adaptive hierarchical cuboid abstractions of {3D} shape
  collections.
\newblock \emph{ACM Transactions on Graphics (TOG)}, 38\penalty0 (6):\penalty0
  1--13, 2019.

\bibitem[Tulsiani et~al.(2017)Tulsiani, Su, Guibas, Efros, and
  Malik]{tulsiani2017learning}
Shubham Tulsiani, Hao Su, Leonidas~J Guibas, Alexei~A Efros, and Jitendra
  Malik.
\newblock Learning shape abstractions by assembling volumetric primitives.
\newblock In \emph{Proceedings of the IEEE Conference on Computer Vision and
  Pattern Recognition}, pages 2635--2643, 2017.

\bibitem[Vu et~al.(2022)Vu, Kim, Luu, Nguyen, and Yoo]{Vu_2022_CVPR}
Thang Vu, Kookhoi Kim, Tung~M. Luu, Thanh Nguyen, and Chang~D. Yoo.
\newblock Softgroup for {3D} instance segmentation on point clouds.
\newblock In \emph{Proceedings of the IEEE/CVF Conference on Computer Vision
  and Pattern Recognition (CVPR)}, pages 2708--2717, 2022.

\bibitem[Wald et~al.(2020)Wald, Dhamo, Navab, and Tombari]{wald2020learning}
Johanna Wald, Helisa Dhamo, Nassir Navab, and Federico Tombari.
\newblock Learning {3D} semantic scene graphs from {3D} indoor reconstructions.
\newblock In \emph{Proceedings of the IEEE/CVF Conference on Computer Vision
  and Pattern Recognition}, pages 3961--3970, 2020.

\bibitem[Wang and Guo(2019)]{wang2019efficientplane}
Chao Wang and Xiaohu Guo.
\newblock Efficient plane-based optimization of geometry and texture for indoor
  rgb-d reconstruction.
\newblock In \emph{Proceedings of the IEEE/CVF Conference on Computer Vision
  and Pattern Recognition Workshops}, pages 49--53, 2019.

\bibitem[Wu et~al.(2022)Wu, Liu, Ruan, and Chirikjian]{wu2022primitive}
Yuwei Wu, Weixiao Liu, Sipu Ruan, and Gregory~S Chirikjian.
\newblock Primitive-based shape abstraction via nonparametric bayesian
  inference.
\newblock In \emph{European Conference on Computer Vision}, pages 479--495.
  Springer, 2022.

\bibitem[Yan et~al.(2021)Yan, Yang, Ma, Huang, Vouga, and Huang]{yan2021hpnet}
Siming Yan, Zhenpei Yang, Chongyang Ma, Haibin Huang, Etienne Vouga, and Qixing
  Huang.
\newblock Hpnet: Deep primitive segmentation using hybrid representations.
\newblock In \emph{Proceedings of the IEEE/CVF International Conference on
  Computer Vision}, pages 2753--2762, 2021.

\bibitem[Yang and Chen(2021)]{yang2021unsupervised}
Kaizhi Yang and Xuejin Chen.
\newblock Unsupervised learning for cuboid shape abstraction via joint
  segmentation from point clouds.
\newblock \emph{ACM Transactions on Graphics (TOG)}, 40\penalty0 (4):\penalty0
  1--11, 2021.

\bibitem[Zhou et~al.(2019)Zhou, Qi, Zhai, Sun, Chen, Wei, and
  Ma]{zhou2019wirefrmae}
Yichao Zhou, Haozhi Qi, Yuexiang Zhai, Qi Sun, Zhili Chen, Li-Yi Wei, and Yi
  Ma.
\newblock Learning to reconstruct {3D} manhattan wireframes from a single
  image.
\newblock In \emph{Proceedings of the IEEE/CVF International Conference on
  Computer Vision}, pages 7698--7707, 2019.

\end{thebibliography}
}

\clearpage
\appendix
\section*{Supplementary Material}
In this supplemental material, we provide detailed explanations of the preprocessing steps and score function used in our rectilinear mesh reconstruction (\Sec{suprecon}), implementation details of transformer network and GCN (\Sec{supimp}), and the synthetic dataset generation process (\Sec{supsynthetic}). 
We also present additional qualitative and quantitative results to demonstrate the effectiveness of our method (\Sec{supeval}). 

\section{Rectilinear Mesh Reconstruction Details}
\label{sec:suprecon}
\subsection{Preprocessing Step}
We details the preprocessing mentioned in Sec. 3.2 of the main paper, which computes the rotation matrix.
A polycuboid instance consists of a set of detected faces, where each face is represented as a set of points with an inferred face label. 
Each face label corresponds to one of six global axes (e.g., ±x, ±y, ±z). 
We utilize this information to compute the rotation matrix.
We first fit a plane to each detected face to obtain its normal vector. 
We then compute the angular differences between the plane normals and the corresponding global axes derived from the inferred face labels. Finally, we determine the rotation matrix that minimizes these angle differences using a least-square approach. 

Using the optimized rotation matrix, we rotate the polycuboid instance to be aligned with the global frame. After completing the rectilinear mesh reconstruction process, we transform the reconstructed polycuboid mesh back to its original coordinate by applying the inverse of the rotation matrix.

\subsection{Score Function}

In Sec. 3.2 of the main paper, to reconstruct a polycuboid instance, we select a set of 3D boxes from a 3D non-uniform grid using a heuristic score function.
Each box $b$ in the grid is represented as a polygonal mesh of a cuboid parameterized by its center and extents, and the score function for $b$ is defined as:
\begin{equation}
S(b) = \sum_{n=1}^{6} sgn(f_n, P_n) \cdot w(f_n, P_n),
\end{equation}
where $f_n$ represents each face of the box $b$, and $P_n$ is the set of points from detected faces located within an 0.05m of $f_n$.  
The sign function $sgn(f_n, P_n)$ returns $1$ if the cuboid face label of $f_n$ matches the most frequent face label in $P_n$ and $-1$ otherwise. The weight factor $w(f_n, P_n)$ mitigates the impacts of partially detected faces due to incomplete scan data.
It is defined as the ratio of the estimated surface area of $P_n$ to the area of $f_n$. 
The area of $P_n$ is computed using the area of the 2D bounding box of projected $P_n$ onto the face $f_n$.
A box with a positive score is classified as inside the polycuboid instance, whereas a negative score is classified as outside.


\section{Implementation Details}
\label{sec:supimp}
We use Transformer and Graph convolutional Network (GCN) in our framework. 
This section provides implementation details of these networks with model architectures.

For the face labeling task, we adopt the model named Stratified Transformer~\cite{lai2022stratified}. 
The model first encodes the input point cloud using a hierarchical tokenization strategy, where local features are extracted at multiple levels of granularity. Theses features are then processed by five transformer layers, where each layer refines the point features based on attention mechanism.
Once the point features are fully embedded, point-wise classification and offset regression are performed using two separate heads, each composed of a two-layer MLP. 

For spatial relationship prediction, we adopt the GCN model proposed by Wald et al.~\cite{wald2020learning}, which is designed to learn structured scene graphs from 3D indoor scene data. 
The model first encodes input data using two separate PointNet encoders, one for node features and another for edge features to transform the raw information into higher-dimensional feature representation.
The node and edge features are then fed into a GCN that 
consists of five graph convolution layers and iteratively refines the node and edge features through message passing. Node classification and edge classification are finally performed using two separate heads composed of three-layer MLP. 

\section{Synthetic Dataset Generation}
\label{sec:supsynthetic}
We provide further details on the dataset generation process described in Sec. 4.1 of the main paper, along with visual examples. 

\begin{figure}[t]  
\begin{center}  
     \begin{subfigure}[b]{0.45\columnwidth}
         \includegraphics[width=\textwidth]{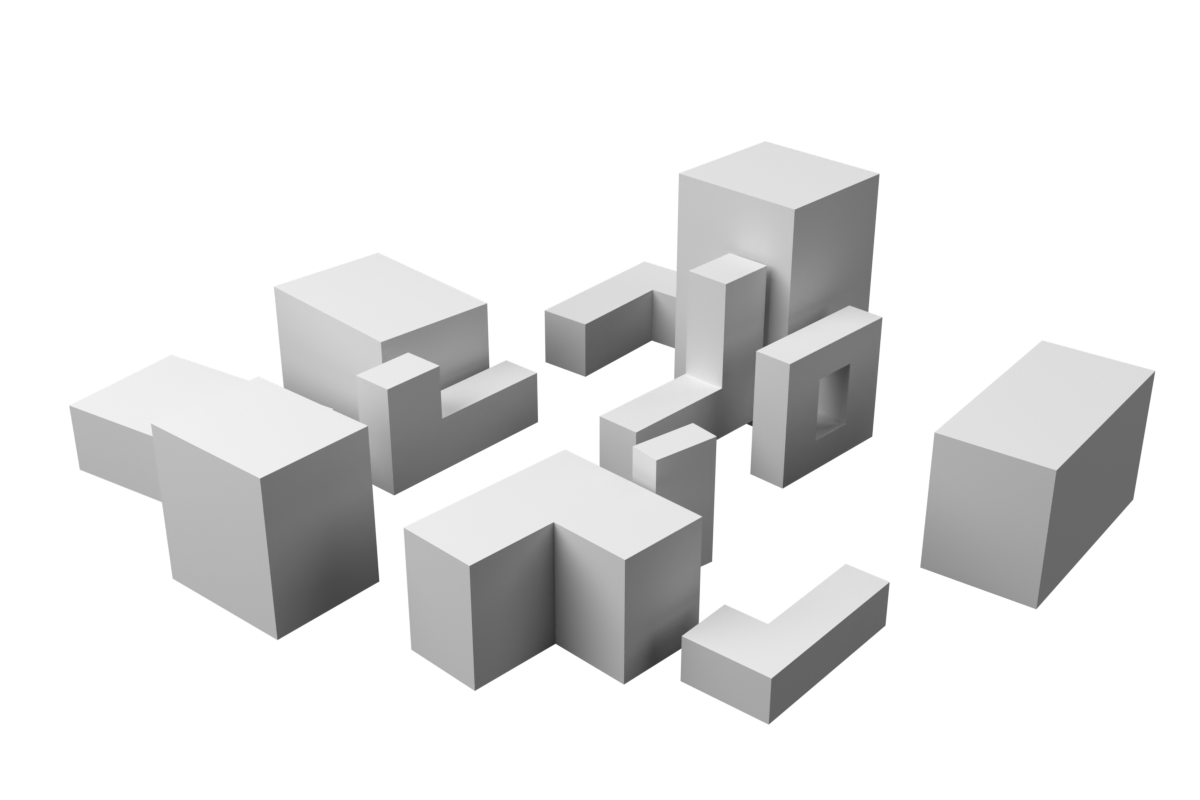}
          \caption{Random configuration}
         \label{fig:synthetic_random}
     \end{subfigure}
     \begin{subfigure}[b]{0.45\columnwidth}
         \includegraphics[width=\textwidth]{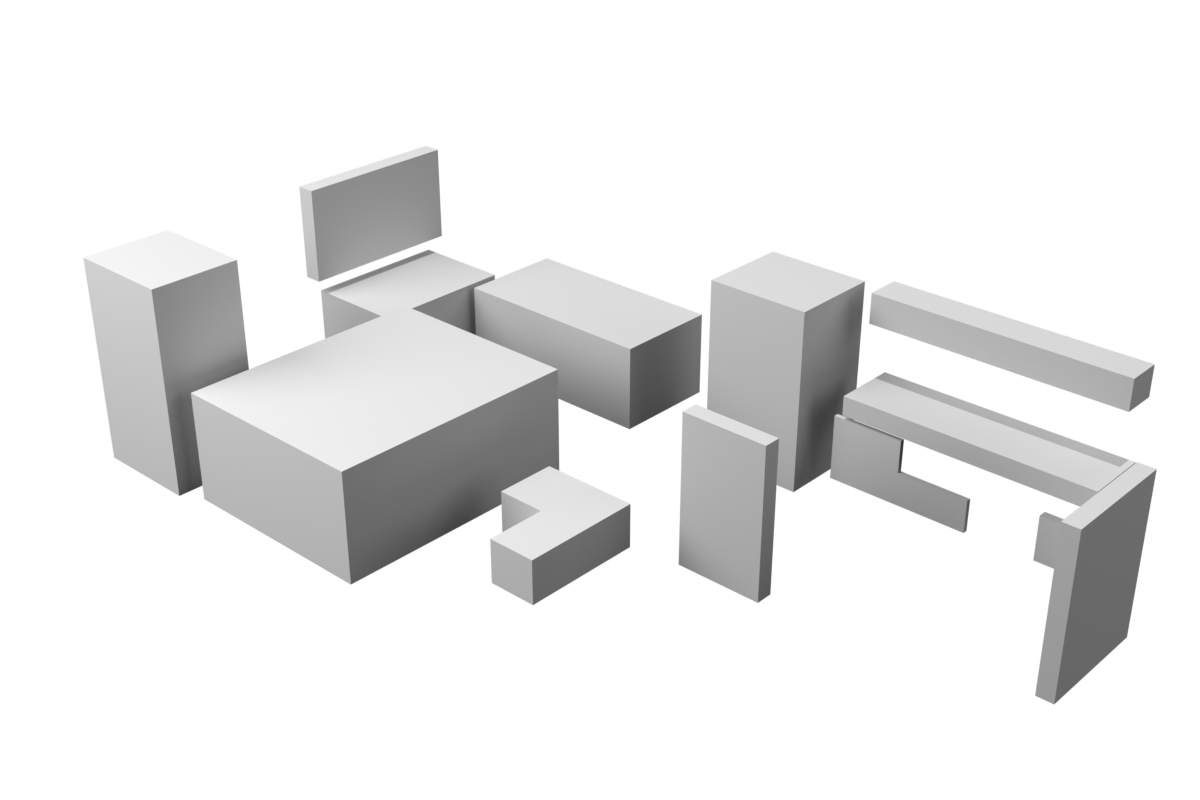}
          \caption{Contextual configuration}
         \label{fig:synthetic_context}
     \end{subfigure}
        \caption{Two types of configurations of polycuboids are used to generate our synthetic dataset. (a) Polycuboids are randomly placed. (b) Polycuboids are positioned to replicate the object configurations derived from the ScanNet dataset. } 
\end{center}
\end{figure}

Each scene is constructed using a combination of polycuboid and cuboid meshes, following two configuration types:
1) Random configuration: Scenes are generated by randomly placing polycuboid and cuboid meshes with varying numbers, scales, and orientations while ensuring that no meshes overlap (\Fig{synthetic_random}).
Each scene contains 5 to 20 meshes whose scales range from 0.4m to 2.4m.
2) Contextual configuration: 
To simulate more realistic configurations, we utilize object bounding boxes from the ScanNet dataset, which are parameterized by their center coordinates and extents.
In these scenes, existing bounding boxes are randomly replaced by polycuboid meshes (\Fig{synthetic_context}).

\begin{figure}[t]  
\begin{center}
  \includegraphics[width=0.8\linewidth, trim=1cm 13cm 8cm 0cm]{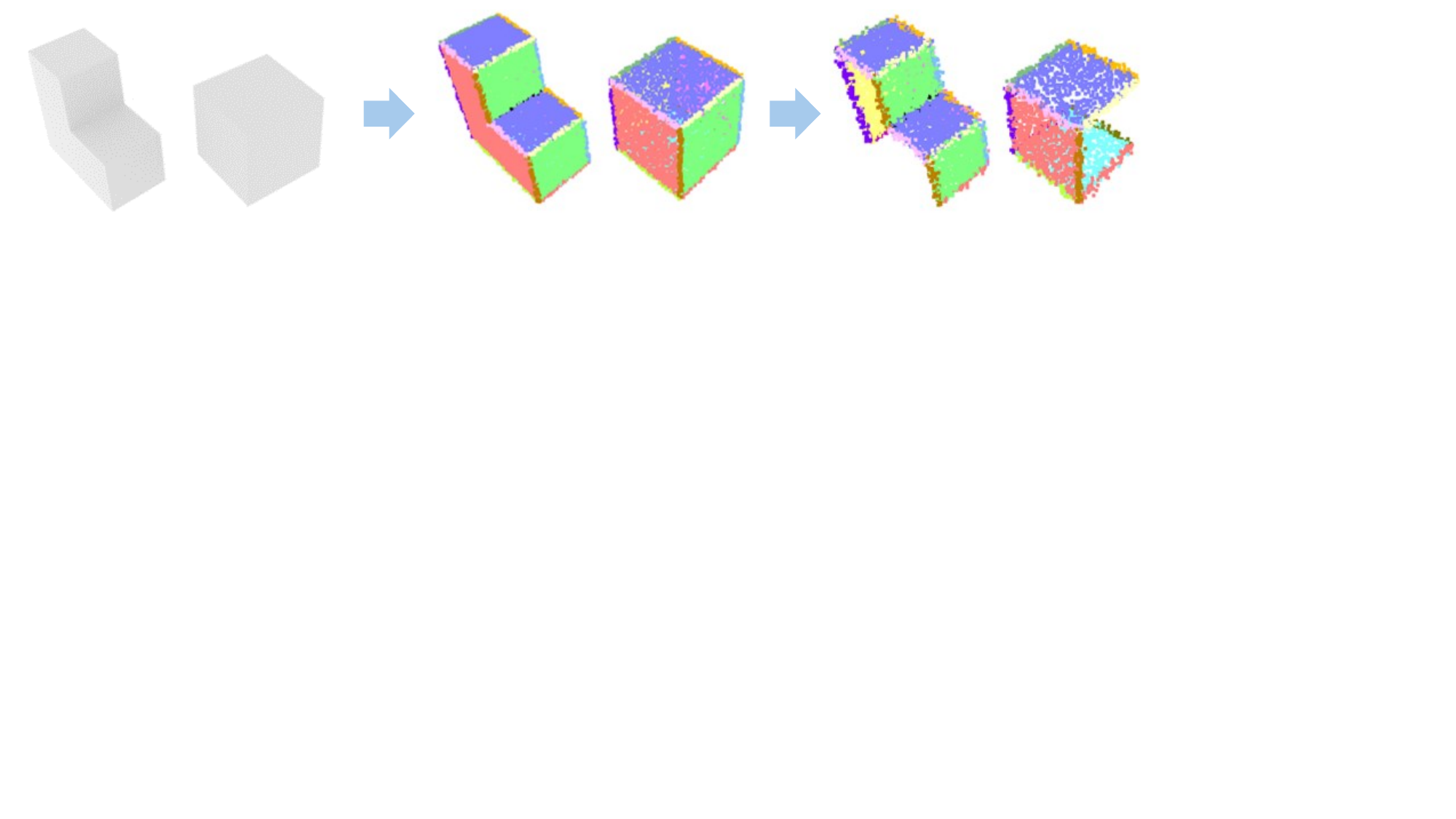}
\end{center} 
\caption{A simplified example of the data generation process. Points are sampled from polycuboid and cuboid meshes, followed by adding noise and holes. }
\label{fig:scanning}
\end{figure}

After composing the scenes, we sample points at 1cm interval on each mesh face. To better simulate real-world scanning conditions, we add Gaussian noise to the sampled points and create holes by randomly removing 0 to 3 faces from each mesh, along with their corresponding sampled points (\Fig{scanning}).


\section{Additional Results}
\label{sec:supeval}
\begin{figure*}[t]  
\begin{center}
  \includegraphics[width=0.8\linewidth, trim=1cm 13cm 8cm 0cm]{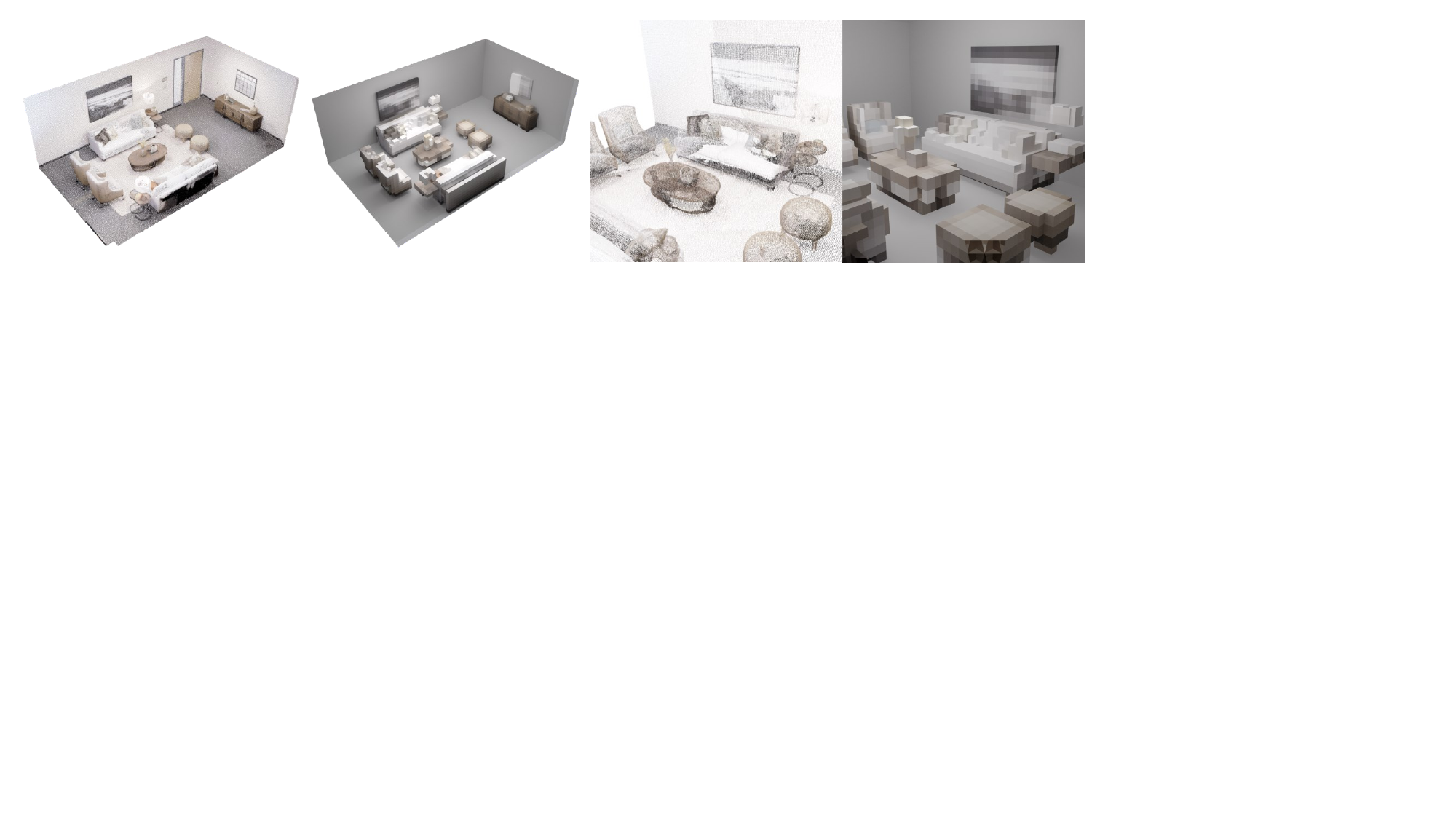}
\end{center} 
\caption{Example of polycuboid abstraction on Replica dataset for `room0'. The two images on the left show the input point cloud and our final result, while the two images on the right provide close-up views. }
\label{fig:supp_replica_sample}
\end{figure*}

In this section, we present additional qualitative and quantitative results to further evaluate our method.
In \Fig{supp_replica_sample}, we illustrate examples of input point clouds alongside our final results for `room0' from the Replica dataset. Despite the noisy input, scene components are plausibly represented with polycuboids.

\begin{figure*}[t]  
\begin{center}
  \includegraphics[width=0.8\linewidth]{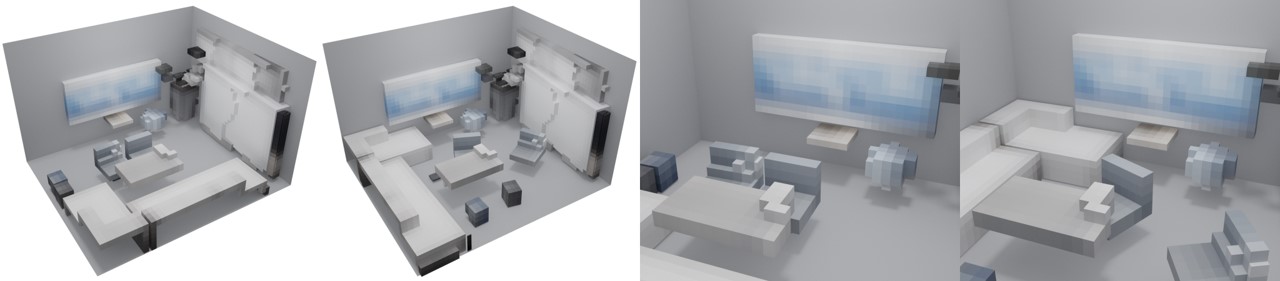}
\end{center} 
\caption{Editing examples of reconstructed polycuboids on Replica dataset for `office 0'. The first and third images show the original arrangements of polycuboids representing the scene, while the second and fourth images show rearranged polycuboids, including an L-shaped sofa and two chairs. }
\label{fig:supp_replica_editing}
\end{figure*}

\begin{figure*}[t]  
\begin{center}
  \includegraphics[width=\linewidth, trim=1cm 13cm 1cm 0cm]{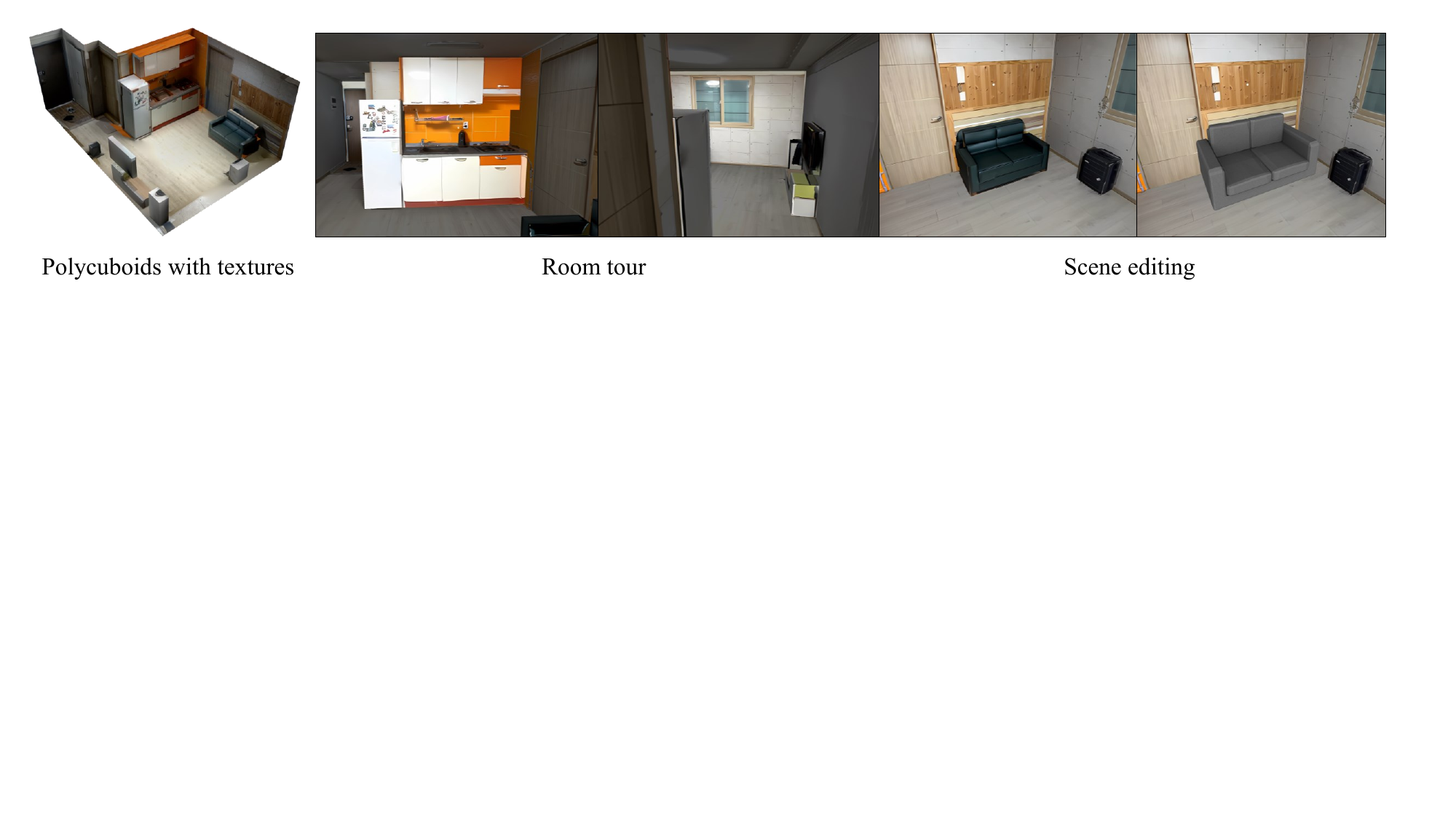}
\end{center} 
\caption{Application examples on the dataset scanned with an iPhone. Our compact polycuboid representation of an indoor scene enables practical applications, such as virtual room tour and scene editing. For scene editing, a grey sofa created by a designer is used to replace the original scanned sofa. }
\label{fig:supp_application}
\end{figure*}

We also demonstrate two applications, scene editing and virtual room tours, in \Figs{supp_replica_editing} and \ref{fig:supp_application}, respectively. These examples highlight the ease of manipulating scene components and the seamless virtual exploration experience, which would be valuable for interior design and AR/VR applications. 

\begin{figure*}[t]
\begin{center}
  \includegraphics[width=0.95\linewidth, trim=0cm 0cm 1cm 10cm]{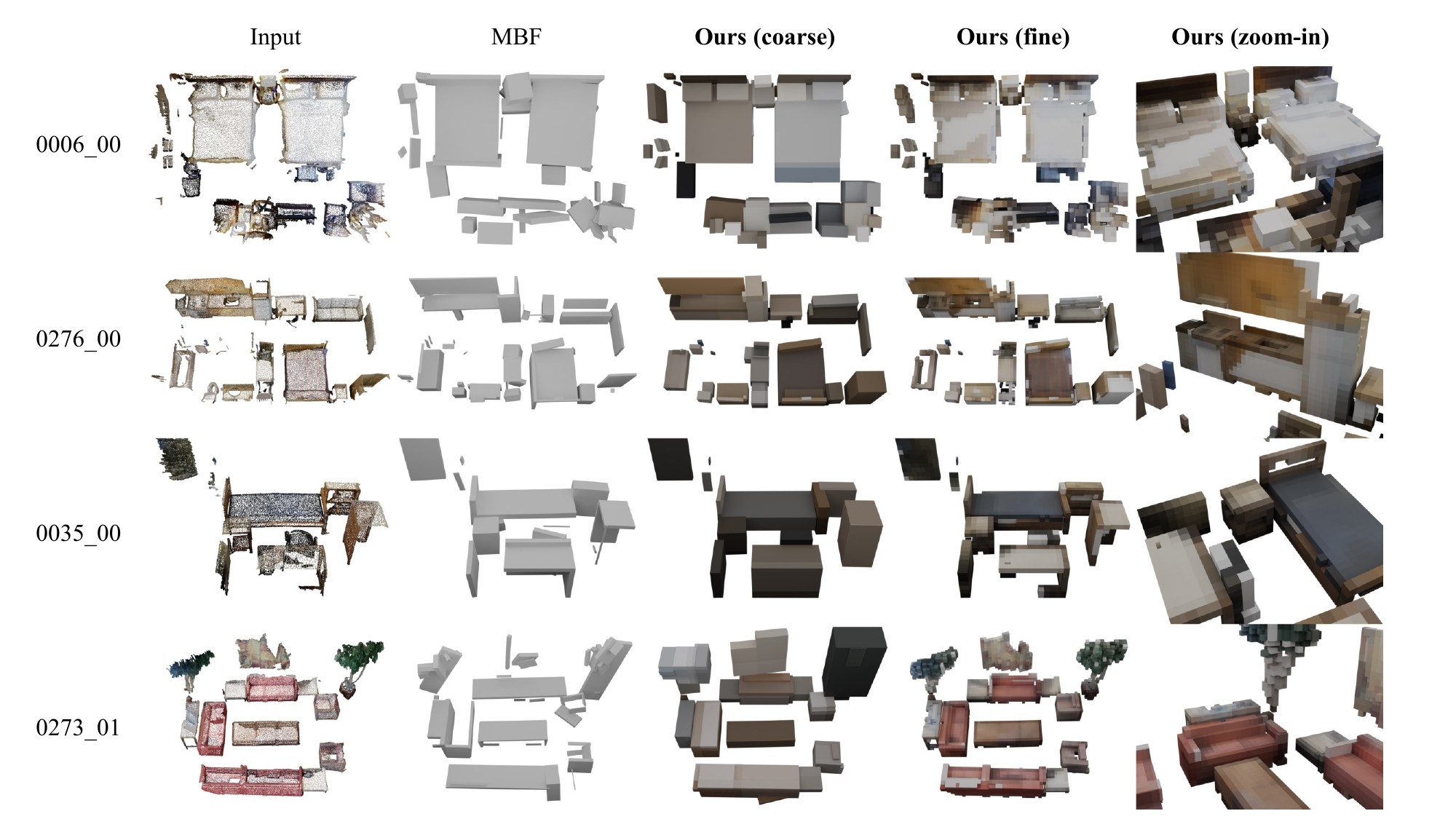}
\end{center}
\caption{Qualitative comparison on four scenes from ScanNet dataset. From left to right, columns show the input point cloud, results from MBF \cite{ramamonjisoa2022monteboxfinder}, results from our method with coarse and fine detail levels, and zoomed-in views of our fine level results. Our method faithfully captures the underlying geometry, even when the inputs are very noisy and incomplete. }
\label{fig:supp_quali_scannet}
\end{figure*}

\begin{figure*}[t]
\begin{center}
\includegraphics[width=0.9\linewidth, trim=1.5cm 1cm 16.2cm 0cm]{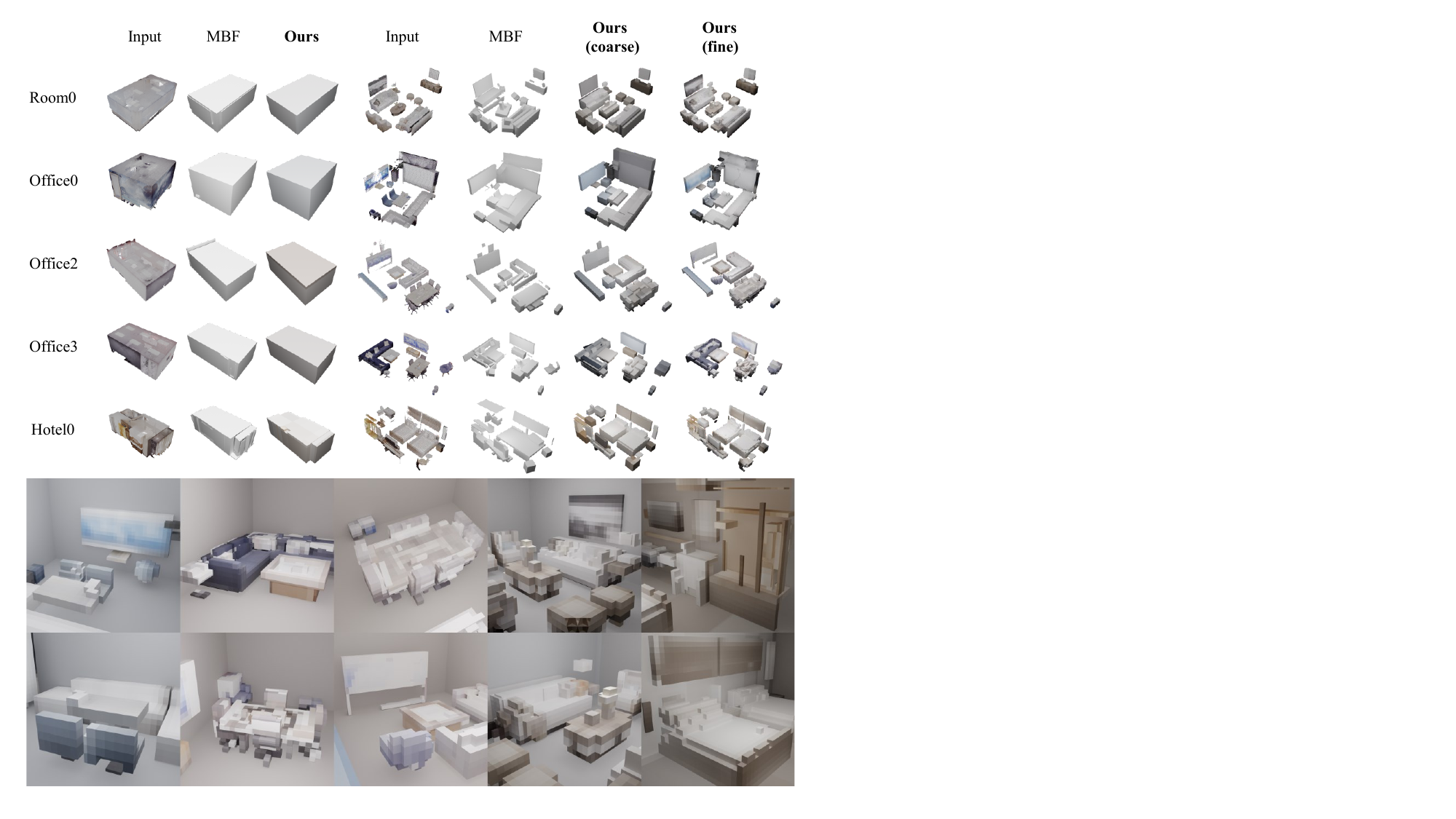}  
\end{center}
   \caption{Qualitative evaluations on five rooms from Replica dataset, showing the separate results for layout and objects. From left to right, columns show the input point cloud for layout, results from MBF \cite{ramamonjisoa2022monteboxfinder}, results from our method, the input point cloud for objects, results from MBF, results from our method with coarse and fine detail levels. The bottom two rows provide zoomed-in views of our results constructed using polycuboids for both layout and objects.
}
\label{fig:supp_quali_replica}
\end{figure*}

Further quantitative and qualitative evaluations are provided on the ScanNet and Replica datasets. Visual results are shown in \Figs{supp_quali_scannet} and \ref{fig:supp_quali_replica}, with corresponding quantitative results in \Tbls{supp_quanti_scannet} and \ref{tbl:supp_quanti_replica}, respectively.

\begin{table*}[t]
\caption{For four scenes from ScanNet dataset, we use Chamfer Distance (CD) to measure geometric discrepancy between input points and points sampled from our output polycuboids. Average: average CD score for four examples. Average$^\ast$: average CD score for all validation data of ScanNet.}
\label{tbl:supp_quanti_scannet}
\begin{center} 
\resizebox{0.72\linewidth}{!}{
\begin{tabular}{lcccccc}
\toprule
  & 0006\_00 &  0035\_00 & 0273\_01 & 0276\_00  & Average & Average$^\ast$ \\
\hline
MBF\cite{ramamonjisoa2022monteboxfinder} &  0.078 & 0.046 & 0.069 & 0.065 & 0.065 & 0.066 \\
\textbf{Ours} & 0.047  & 0.034 & 0.034 & 0.087 & 0.050 & 0.040 \\
\hline
\toprule
\end{tabular} 
}
\end{center} 
\end{table*}

\begin{table*}[t]
\caption{For nine scenes from Replica dataset, we use Chamfer Distance (CD) to measure geometric discrepancy between input points and points sampled from our output polycuboids. We apply our framework to points categorized under `layout' and `non-layout', respectively. }
\label{tbl:supp_quanti_replica}
\begin{center}
\begin{tabular}{lccccccccccc}
\toprule
 & type &  hotel 0 &  office 0   &  office 1 & office 2 & office 3 & office 4 & room 0 & room 1 & room 2 & Average  \\
\hline 
 
MBF\cite{ramamonjisoa2022monteboxfinder} & non-layout &  0.068 & 0.151  &  0.068 & 0.074 &  0.076 &  0.071 & 0.060 & 0.078 & 0.056 &  0.078 \\
\textbf{Ours} & non-layout&  0.040 &  0.037 &  0.051 & 0.040 & 0.033 & 0.036&  0.048&   0.075 &  0.036&  0.044\\
\hline
MBF\cite{ramamonjisoa2022monteboxfinder}&layout & 0.078 & 0.062 &  0.049 &  0.036  &    0.043  &  0.027 & 0.037 &  0.049  &  0.036 &  0.047\\
\textbf{Ours} &layout &  0.131 &  0.090  & 0.073 &  0.042  &  0.116 &  0.045 &  0.064 &  0.082 &  0.084  & 0.081 \\
\hline
\toprule
\end{tabular}
\end{center}
\end{table*}

\end{document}